\documentclass[runningheads]{llncs}

\usepackage[mobile]{eccv}

\usepackage{eccvabbrv}

\usepackage{graphicx}
\usepackage{booktabs}

\usepackage[accsupp]{axessibility}  

\usepackage{hyperref}
\usepackage{multirow} 
\usepackage{pifont}
\usepackage{float} 
\usepackage{placeins} 
\usepackage{afterpage}
\usepackage{changepage}
\usepackage{subcaption} 
\newcommand{\cmark}{\ding{51}}  

\usepackage{colortbl, xcolor}
\usepackage{makecell} 
\usepackage[table]{xcolor}  
\usepackage{booktabs}     
\usepackage{cuted}
\usepackage[hang,flushmargin]{footmisc}
\usepackage{enumitem}
\newcommand{\blue}[1]{\textcolor{blue}{#1}}

\definecolor{lightgray}{gray}{0.92}

\usepackage{orcidlink}
\usepackage{array}
\newcolumntype{C}[1]{>{\centering\arraybackslash}p{#1}}

\definecolor{blue1}{HTML}{4472C4}
\definecolor{orange1}{HTML}{ED7D31}
\definecolor{green1}{HTML}{548235}
\definecolor{red1}{HTML}{C00000}
\usepackage[most]{tcolorbox}      
\usepackage{multirow}

\newtcolorbox{casestudy}[2][]{%
  colback=gray!5,
  colframe=gray!50,
  fonttitle=\bfseries,
  title={#2},
  breakable,
  enhanced,
  left=1em,
  right=1em,
  top=0.8em,
  bottom=0.8em,
  #1
}

\begin{document}

\title{Empowering Long-form Omni-modal Understanding with Robust Audio Perception} 

\titlerunning{ }

\author{Kaiying Yan\inst{1} \and
Luoyi Sun\inst{2,3} \and
Xiao Zhou\inst{1}\and
Weidi Xie\inst{1} 
}

\authorrunning{ }

\institute{SAI, Shanghai Jiao Tong University, shanghai, China \and Zhejiang University, Hangzhou, Zhejiang, china, 
\and
Shanghai AI Lab, shanghai, China}

\maketitle 
\footnotetext[2]{Corresponding author}

\begin{abstract}
Recent advances in large-scale multimodal models have driven remarkable progress in vision-language tasks; however, comprehensive omni-modal understanding remains under-explored, largely due to the scarcity of datasets with rich, explicitly aligned auditory cues. To bridge this gap, we present \textbf{\texttt{AVDC}}~(\textbf{\texttt{A}}udio-\textbf{\texttt{V}}isual \textbf{\texttt{D}}ecoupled \textbf{\texttt{C}}aptions), a large-scale dataset designed to disentangle visual and auditory semantics. Specifically, we propose an automated pipeline that leverages off-the-shelf models to annotate videos with tripartite captions: visual-only (V), audio-only (A), and joint audio-visual (AV). This decoupled structure explicitly captures both modality-specific nuances and complex cross-modal interactions. Building upon this, we introduce \textbf{\texttt{AVDC-QA-CoT}}, a Chain-of-Thought augmented question-answering dataset to foster audio-visual reasoning. To fully exploit these resources, we employ a two-stage training paradigm: omni-modal caption generation pre-training on \textbf{\texttt{AVDC}}, followed by instruction tuning on \textbf{\texttt{AVDC-QA-CoT}}. Extensive experiments across diverse downstream tasks, spanning video captioning, audio-centric analysis, and omni-modal benchmarks, demonstrate consistent and significant performance gains, showing the efficacy of our proposed datasets and training strategy in advancing omni-modal perception. Code and dataset are related on \url{https://radiant0726.github.io/AVDC-web/}.
\end{abstract}
   

\section{Introduction}
\label{sec:intro}

\begin{figure}[t]
    \centering
    \includegraphics[width=\linewidth]{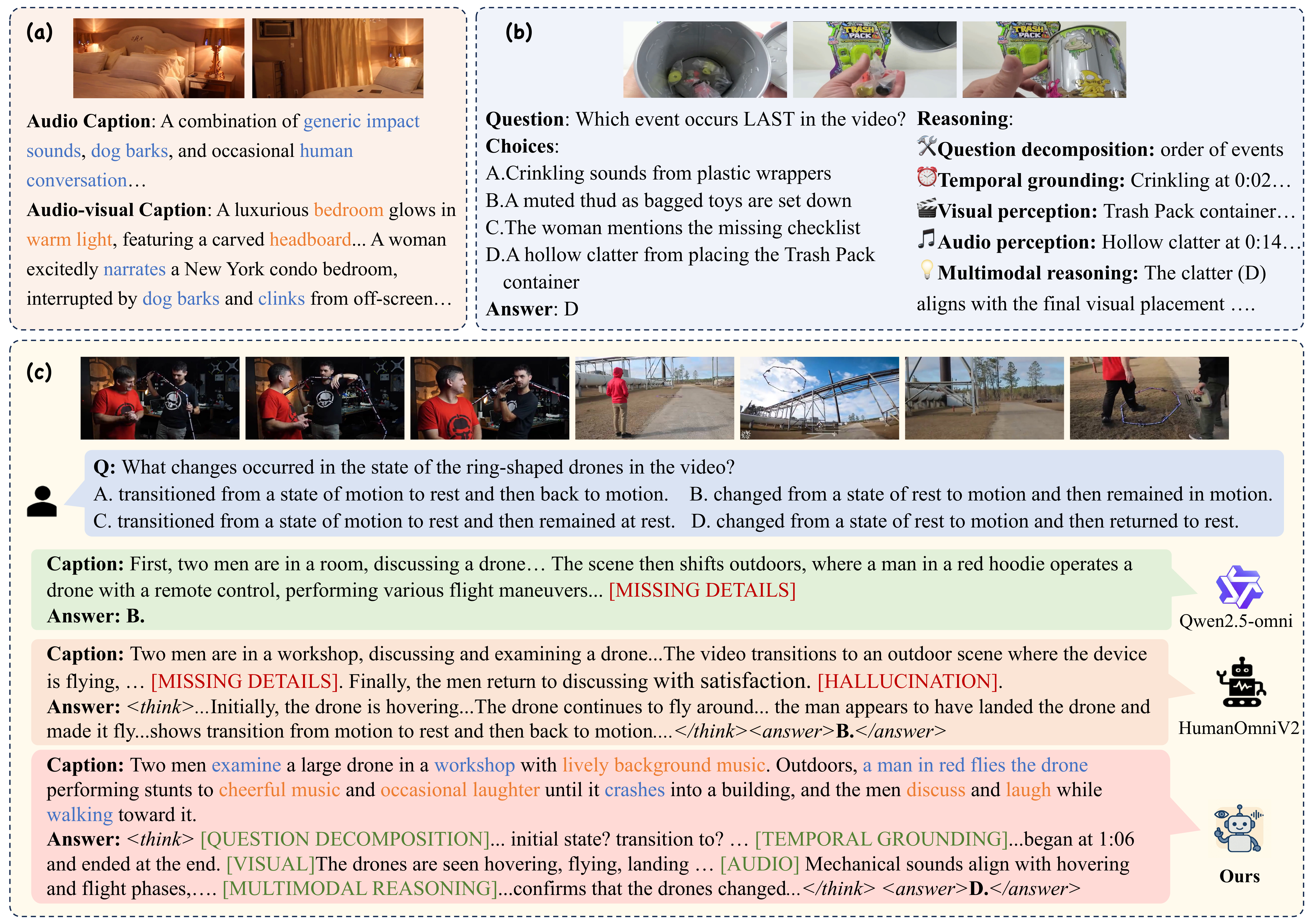}
    \vspace{-2.0em}
    \caption{\textbf{Dataset and qualitative results.} 
\textbf{\texttt{(a)}} \textbf{\texttt{AVDC}} dataset. An audio-visual caption dataset with balanced information across modalities, 
\textbf{\texttt{(b)}} \textbf{\texttt{AVDC-QA-CoT}} dataset. It include QA pairs with chain-of-thought reasoning. \textbf{\texttt{(c)}} QA results visualization. The comparison demonstrates that our model generates more detailed and accurate captions with fewer missing or hallucinated events, and exhibits more coherent omni-modal reasoning. 
Vision-related and audio-related contents are highlighted in \textcolor{orange1}{orange} and \textcolor{blue1}{blue}, respectively. The parts highlighted in \textcolor{red1}{[red]} and \textcolor{green1}{[green]} indicate caption problems and answer reasoning steps, respectively.}
    \label{fig:teaser}
    \vspace{-2.0em}
\end{figure}


Perception is inherently omni-modal; we rely on acoustic cues not only to complement visual information but also to perceive events beyond our field of view. While recent Multimodal Large Language Models (MLLMs) have achieved remarkable success in video understanding~\cite{fu2025video, li2024mvbench}, they remain predominantly vision-centric~\cite{caffagni2024revolution, nguyen2024video}. This visual bias severely hampers performance in scenarios where audio is the primary information source, such as distinguishing a running blender from a dishwasher, localizing an off-screen siren, or inferring intent through prosody. In such cases, models frequently hallucinate based on visual priors or rely on textual shortcuts. We posit that this limitation stems not merely from architectural constraints, but from a systemic bias in dataset design: contemporary training corpora lack the dense, independent auditory information required to force models to truly \textit{listen} rather than just \textit{look}.

Specifically, existing multimodal datasets exhibit two major deficiencies. 
First, audio supervision is often dominated by speech transcripts~\cite{wu2025galaxy, huynh2025svla}, neglecting environmental sounds and complex acoustic scene dynamics. 
Second, even when sound events are annotated, they are typically synchronized with visible objects~\cite{cheng2024avset, lee2021acav100m}. This `visual redundancy' allows models to predict auditory events solely from visual features, rendering the audio modality largely auxiliary. The scarcity of non-speech, off-screen sound events prevents models from developing robust cross-modal reasoning capabilities.

In this paper, we introduce an automated data construction pipeline designed to integrate dense visual and auditory information, 
yielding two complementary datasets: \textbf{\texttt{AVDC}}, a dense audio-visual decoupled captioning dataset, and \textbf{\texttt{AVDC-QA-CoT}}, an omni-modal QA corpus enriched with Chain-of-Thought (CoT) rationales. By leveraging specialized audio models and large language models, we generate three aligned captions per video: audio-only, visual-only, and joint audio-visual. As illustrated in Figure~\ref{fig:teaser}, this tripartite structure explicitly captures non-speech events and intricate cross-modal interactions. 
These captions serve as the foundation for synthesizing open-ended and multiple-choice QA pairs with chain-of-thought rationales for modality-specific and complex reasoning tasks. Consequently, we curate a collection of 10,000 long-form videos, each paired with dense tripartite descriptions and QA instances. This facilitates supervision that is simultaneously context-consistent, fine-grained, and explicitly decoupled across modalities. On average, each video contains 5.5 sound events, with around 42\% of them attributed to invisible sound sources, representing a substantial increase compared to existing datasets.

To validate the efficacy of our proposed datasets, we conduct comprehensive experiments across diverse omni-modal understanding tasks. At training time, we adopt a two-stage paradigm that prioritizes auditory competence and mitigates vision-text shortcuts. First, we post-train \textit{Qwen-2.5-Omni-7B-Thinker} on the \textbf{\texttt{AVDC}} dataset for caption generation, to establish robust acoustic event comprehension. Second, we perform instruction tuning on a diverse mixture of \textbf{\texttt{AVDC-QA-CoT}}, and standard vision question answering~(VQA)~\cite{feng2025video, Maaz2023VideoChatGPT}, and audio question answering~(AQA) datasets~\cite{ghosh2025audio,goel2025audio,ghosh2024gama}, thereby improving omni-modal reasoning while preserving audio sensitivity. 

We evaluated on a comprehensive suite of benchmarks. Specifically, we first introduce \textbf{\texttt{AVDC-test}}, a split specifically curated to quantify performance on visible versus invisible sound events. Furthermore, we evaluate on omni-modal captioning ({\em e.g.}, Video-SALMONN-2~\cite{tang2025video}, UGC-VideoCap~\cite{wu2025ugc}), general omni-modal understanding ({\em e.g.}, WorldSense~\cite{hong2025worldsense}, Omnibench~\cite{li2024omnibench}), and specialized audio understanding tasks (MMAU~\cite{sakshi2024mmau}). The results demonstrate that our data-centric approach significantly narrows the gap between visual and auditory comprehension, yielding state-of-the-art performance on audio-centric and cross-modal reasoning tasks.

\begin{figure*}[!t]
    \centering
    \includegraphics[width=\linewidth]{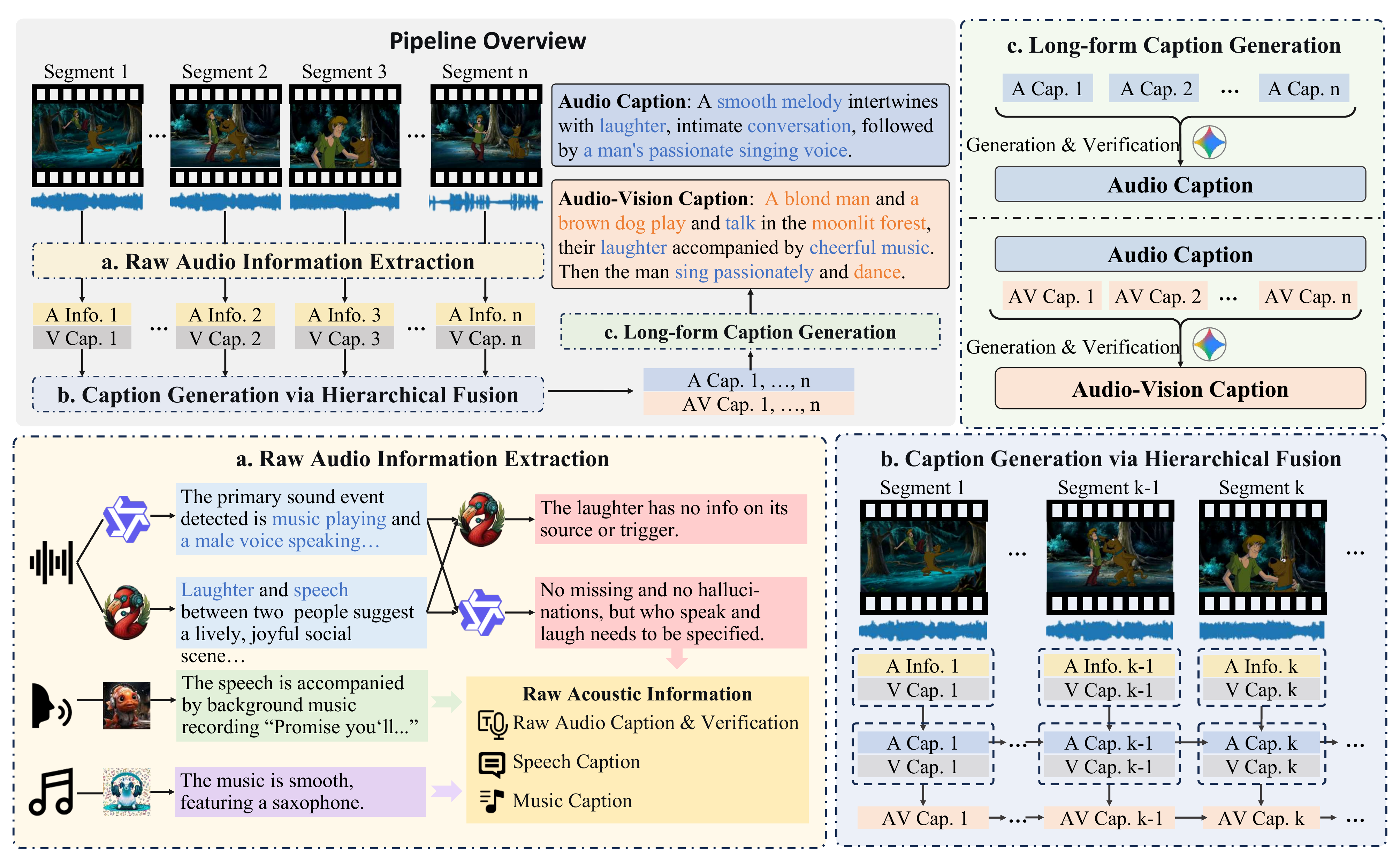}
    \vspace{-0.7cm}
    \caption{\textbf{Automatic pipeline for the audio-visual decoupled caption generation.} \textbf{I.} Multiple audio and language models collaborate to extract raw auditory information, including general audio captions, music and speech descriptions, and verification results. 
    \textbf{II.} The extracted audio information is temporally aligned and integrated with visual content to produce segment-level audio and audio-visual captions in sequential order, ensuring temporal continuity. 
    \textbf{III.} The long-form caption generation stage summarizes all segment-level captions into a coherent global video caption, while multimodal verification and correction refine accuracy and consistency. Vision- and audio-related content is highlighted in \textcolor{orange1}{orange} and \textcolor{blue1}{blue}, respectively. Abbreviations: Cap.~(Caption), Info.~(Information).}
    \label{fig:pipeline}
    \vspace{-0.5cm}
\end{figure*}

\section{Dataset Construction}
This section details the automated pipeline to construct data that captures the fine-grained, temporal evolution of visual and auditory dynamics. Specifically, we introduce two novel datasets: (i) \textbf{\texttt{AVDC}}, a dense, decoupled audio-visual captioning corpus, and (ii) \textbf{\texttt{AVDC-QA-CoT}}, an instruction-tuning corpus designed for complex question answering equipped with Chain-of-Thought (CoT) rationales. 
Our data generation framework operates in two distinct stages, namely, \textit{omni-captioning} and \textit{omni-instruction generation}. Detailed prompt templates and system configurations are provided in the supplementary materials.

\subsection{Audio-Visual Decoupled Caption Dataset}
\label{Omni Captioning}

Audio carries information that vision often misses, for example, sound events, speech, and music can complement or even replace visual cues, 
it is therefore crucial for omni-modal understanding. We introduce a high-quality dataset of \textbf{long-form, dense audio-visual captions} with \textbf{explicit timestamps}. As shown in Figure~\ref{fig:pipeline}, we first apply a suite of audio models to extract cues from the soundtrack, then use large language models to fuse audio-visual evidence for caption generation, filtering, and refinement, ensuring high-fidelity annotations.

\vspace{3pt} \noindent \textbf{Pre-processing.} 
We collect videos from the open-source video-captioning datasets ShareGPT4Video~\cite{chen2024sharegpt4video} and Vript~\cite{yang2024vript}, spanning diverse domains ({\em e.g.}, news, sports, and travel). 
To ensure acoustic diversity, we apply ATST-F~\cite{schmid2025effective} to categorize audio segments as speech, music, or environmental sounds, and only retain videos that contain at least three distinct environmental sound events (excluding speech and music). We also require these events to be sufficiently dense: their total duration must exceed 2\,s and account for more than 10\% of the video.
For temporal segmentation, we detect transitions in both modalities: visual scene cuts are identified using PySceneDetect\footnote{\url{https://github.com/Breakthrough/PySceneDetect}}, and audio boundaries are placed at the onset or offset of detected sound events. We place a segment boundary whenever a change occurs in either modality, and then merge or split adjacent clips to keep segment durations within $[2, 30]$ seconds, yields $n$ segments, $\mathcal{S} = \{s_1, s_2,\dots, s_n\}$, that that capture multimodal semantic transitions while balancing content sufficiency and event granularity.

\vspace{3pt}\noindent \textbf{Audio information extraction.}
As shown in Figure~\ref{fig:pipeline}a, to capture the full spectrum of auditory cues, for each segment, we extract multi-faceted raw audio information, including sound events, speech, and music, using off-the-shelf audio-language models. We employ Qwen2-Audio~\cite{chu2024qwen2} and Audio-Flamingo 2~\cite{ghosh2025audio} for general audio understanding, SALMONN~\cite{tang2023salmonn} for speech, and MU-LLaMA~\cite{liu2024music} for music analysis. To enhance reliability, the two general audio-language models cross-verify each other’s predictions, and provide binary judgments and brief rationales. These verification notes are retained as auxiliary supervisory signals for later training stage: 
\begin{align}
\mathcal{I}^a_k = \phi_{\text{Audio-LLMs}}(s^a_k),
\end{align}
where $s^a_k$ refers to the audio stream of video segment $s_k$, 
$\mathcal{I}^a_k$ consists of the audio captions generated by the models and corresponding verification rationales.

\vspace{3pt}\noindent \textbf{Caption generation via hierarchical fusion.}
We adopt a hierarchical fusion strategy to produce captions that are temporally coherent and cross-modally consistent. For each segment, generated captions incorporate information from the current and immediately preceding segments to maintain context. Specifically, as shown in Figure~\ref{fig:pipeline}b, for the $k$-th segment, the local captions are generated via large language model~(LLM) by sequentially integrating audio, visual, and contextual information:
\begin{align}
\mathcal{C}^a_k = \phi_{\text{LLM}}(\mathcal{I}^a_k,\mathcal{C}^a_{k-1},\mathcal{C}^v_k), \text{ \hspace{5pt}} 
\mathcal{C}_k  = \phi_{\text{LLM}}(\mathcal{C}^a_k, \mathcal{C}^v_k),
\end{align}
where $\mathcal{C}^\text{a}_k, \mathcal{C}^\text{v}_k, \mathcal{C}_k$ refer to the audio, vision, audio-visual caption for $k$-th video segments. 
For the first segment that lacks preceding context, the model generates $\mathcal{C}^a_1$ and $\mathcal{C}_1$ using only its own audio and visual inputs.

\vspace{3pt}\noindent \textbf{Long-form caption generation and verification.}
As illustrated in Figure~\ref{fig:pipeline}c, we concatenate all segment-level captions in temporal order and perform global refinement using LLM to produce the final video-level descriptions:
\begin{align}
\mathcal{C}^{a}_g = \phi_{\text{LLM}}(\mathcal{C}^{a}_1, \dots, \mathcal{C}^{a}_n), \text{ \hspace{5pt}}
\mathcal{C}_g = \phi_{\text{LLM}}(\mathcal{C}^{a}_g, \mathcal{C}_1, \dots, \mathcal{C}_n),
\end{align}
where $\mathcal{C}^{a}_g$ and $\mathcal{C}_g$ refer to the global audio caption and global audio-visual caption for videos, respectively. 

To ensure high-quality data, we apply an iterative verification pipeline for both video-level and segment-level captions using Gemini‑2.5‑Flash~\cite{comanici2025gemini}. 
For audio captions, we design iterative verification-correction rounds where the model analyzes both audio and visual modalities to detect inconsistencies, such as mismatches, omissions, hallucinations, redundancies, and ambiguities. Minimal edits are applied to preserve correct content, and only captions achieving a quality score of 4 or higher (out of 5) are retained. We adopt an identical verification scheme for audio-visual captions. This procedure effectively corrects recoverable errors and filters out low-quality samples, enhancing the alignment between textual descriptions and multimodal content. The final verified audio and audio-visual caption sets are denoted as $\hat{\mathcal{C}}^{a} = \{\hat{\mathcal{C}}^{a}_g, \hat{\mathcal{C}}^{a}_1, \dots, \hat{\mathcal{C}}^{a}_n\}$ and $\hat{\mathcal{C}} = \{\hat{\mathcal{C}}_g, \hat{\mathcal{C}}_1, \dots, \hat{\mathcal{C}}_n\}$.

\begin{table}[t]
\centering
\caption{Comparison of audio-visual caption dataset. \#V, $L_{\text{V}}$, and $L_{\text{C}}$ represent the number of videos, the average length of each video, and the average token number of each caption, respectively. \#Visible and \#Invisible denote the average number of visible and invisible sound events in each video. 
Abbreviations: Snd. (Sound), Spch. (Speech), Mus. (Music). }
\vspace{-0.3cm}
\scalebox{0.8}{
\begin{tabular}{p{3.0cm}|C{1.0cm}C{1.0cm}C{1.0cm}|C{3.3cm}|C{1.5cm}C{1.5cm}}
\toprule
  &       &        &      &       & \multicolumn{2}{c}{Sound Events} \\
    
\multirow{-2}{*}{Dataset} & \multirow{-2}{*}{\#V} & \multirow{-2}{*}{$L_{\text{V}}$} & \multirow{-2}{*}{$L_{\text{C}}$} & \multirow{-2}{*}{Acoustic Type}      & \#Visible & \#Invisible          \\ 

\midrule

VALOR-1M~\cite{chen2023valor}    & 1.18M & 10.0     & 16.4      & Snd. + Spch. + Mus.         & 1.2& 0.3 \\
VAST-27M~\cite{chen2023vast}   & 27M  & 13.4   & 32.4      & Snd. + Spch. + Mus.         & 1.1& 0.3 \\
AVCaps~\cite{sudarsanam2025avcaps}     & 2,176  & 50.5  & 11.6     & Snd. + Mus.        & 2.0& 0.2 \\
UGC-VideoCap~\cite{wu2025ugc} & 1K    & 23.9  & 184.4   & Snd. + Spch. + Mus.         & 1.1& 0.5 \\
\rowcolor{lightgray}AVDC   &10K   & 30.1  & 258.1      & Snd. + Spch. + Mus.  & \textbf{3.9}& \textbf{2.3} \\

\bottomrule

\end{tabular}
}
\label{tab:stats}
\vspace{-0.5cm}
\end{table}

\vspace{3pt}\noindent \textbf{Dataset statistics.} 
The \textbf{\texttt{AVDC}} dataset comprises 10,000 videos, 
each averaging around 30s in duration and 258 caption words, 
as summarized in Table~\ref{tab:stats}. We randomly sample 1,000 videos from each dataset and use GPT‑4o to detect and categorize sound events as visible or invisible. In \textbf{\texttt{AVDC}}, an average of 3.9 events are visually observable and 2.3 are acoustically present but visually absent, resulting in 37\% invisible events. Compared with existing audio‑visual caption datasets, our proposed dataset features the most sound events per video and the highest proportion of invisible ones, encompassing a mixture of sounds, music, and speech. More details are provided in the supplementary materials.

\vspace{3pt}\noindent \textbf{Manual evaluation.} 
We employ multiple models to generate accurate and comprehensive captions, and subsequently evaluate their reliability. Notably, the two general-purpose audioLLMs and language model APIs play a primary role in the automatic pipeline, while a music- and speech-specialized model is incorporated as a supplementary reference.
To \textbf{measure inter-model consistency between two general audioLLMs on event-level accuracy}, 
we manually annotate sound and visual events for 300 random samples. 
For the two general AudioLLMs used, Qwen2-Audio and Audio-Flamingo 2, we observe low inter-model consistency (Cohen’s $\kappa = 0.134$), indicating substantial model-specific biases. Individually, they achieve F1 scores of 0.467 and 0.449 against the ground truth, respectively. However, merging their predictions using DeepSeek increases the F1 score to 0.671, which further improves to 0.752 after verification with Gemini-2.5-Flash. Incorporating visual cues further raises the sound-event F1 to 0.786, reaching 0.885 after a second verification step. Similarly, the vision-event F1 improves from 0.842 to 0.921 after verification by Gemini-2.5-Flash, demonstrating the effectiveness of our multi-stage refinement pipeline.



\subsection{AVDC-QA-CoT Dataset}\label{sec:AVDC-QA-CoT dataset}
Building upon our omni-captioned video dataset, 
we construct \textbf{\texttt{AVDC-QA-CoT}} to enable audio-visual joint reasoning for video understanding,
that covers recognition, reasoning, temporal localization, and cross-modal consistency, including both open-ended and multiple-choice questions.

\vspace{3pt} \noindent \textbf{Generating question-answering with CoTs.}
Each video is annotated with global and segment-level captions, forming an audio-visual caption set $\hat{\mathcal{C}} = \{\hat{\mathcal{C}}_g, \hat{\mathcal{C}}_1, \dots, \hat{\mathcal{C}}_n\}$. We randomly sample a caption $\hat{\mathcal{C}}_i \in \hat{\mathcal{C}}$ as input and prompt DeepSeek to produce question-answer (QA) pairs, augmented with structured chain-of-thought (CoT) reasoning:
\begin{align}
\{q, a, r\} = \phi_{\text{LLM}}(\hat{\mathcal{C}}_i),
\end{align}
where $q, a, r$ denote the question, the output answer, and the reasoning rationale, respectively.
To promote transparent and evidence-grounded reasoning, we instruct DeepSeek to generate structured rationales through a four‑step process:
(i) decompose the question into semantic components~({\em e.g.}, entities, attributes, relations);
(ii) temporally ground reasoning by locating relevant caption segments;
(iii) extract salient modality‑specific cues-visual~({\em e.g.}, objects, actions, scenes) and auditory~({\em e.g.}, speech, sound events, ambient sounds);
(iv) integrate multimodal evidence to produce a coherent, contextually grounded answer.

\vspace{3pt} \noindent \textbf{Data filtering.}
To ensure that the QA set genuinely requires multimodal reasoning rather than relying on superficial textual shortcuts, we filter all candidate samples using Qwen3‑4B‑Instruct~\cite{yang2025qwen3}. For multiple-choice questions, we discard any instance that Qwen3 answers correctly based only on questions, without access to videos. For open-ended questions, we remove samples whose text-only predictions obtain a METEOR~\cite{banerjee2005meteor} similarity above 0.2 with the reference answer.

\vspace{3pt} \noindent \textbf{Data statistics.} 
The dataset comprises 10,000 audio-visual caption-derived QA pairs. 
Question types are evenly distributed across core video understanding aspects, {\em e.g.}, objects, actions, temporal, spatial cues, causality, counting, and localization-covering diverse multimodal reasoning skills. Answer formats are balanced, with comparable proportions of multiple-choice and open-ended questions and an average rationale length of about 134 words. More details are provided in the supplementary materials.

\section{Methods}

\begin{figure*}[!t]
\centering
\includegraphics[width=\linewidth]{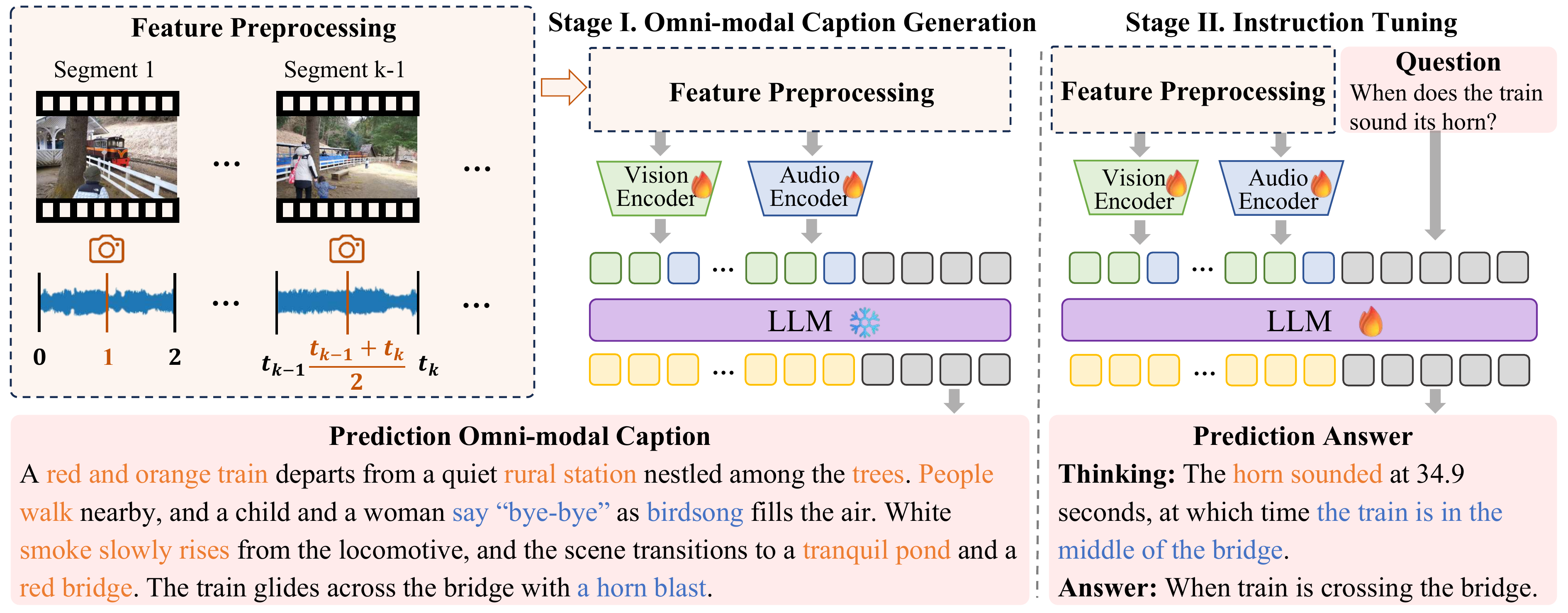}
\vspace{-0.7cm}
\caption{\textbf{Model architecture.} The framework preserves fine-grained audio-visual correspondence via feature preprocessing and employs a two-stage training strategy: \textbf{Stage I.} Omni-modal caption generation training to optimal multimodal representations, and \textbf{Stage II.} Instruction Tuning for QA, supporting both direct answers and chain-of-thought reasoning. Abbreviations: Cap. (Caption), Ans.
(Answer).}
\label{fig:training}
\vspace{-0.3cm}
\end{figure*}

\subsection{Problem Formulation}
Assuming a given video consisting of $m$ frames, denoted as $\mathcal{V} = \{f_1, f_2, \dots, f_m\}$, and its corresponding raw audio waveform $\mathcal{A} = \{w_1, w_2, \dots, w_l\}$.
Along with the question $\mathcal{Q}$, our objective is to train a unified model that can jointly process these multi-modal inputs and answer the question. 
Formally, the model can be seen as a function
\begin{align}
    \pi:(\mathcal{V},\mathcal{A},\mathcal{Q})\mapsto \mathcal{O}
\end{align}
where $\mathcal{O}$ is the output answer to question $\mathcal{Q}$, which can be either a free-text response in open-ended scenarios or an option in a multiple-choice setting.

\subsection{Multimodal Feature Processing}
We adopt a fine-grained, modality-aligned sampling strategy that encodes visual frames and audio segments as temporally synchronized token sequences as unified omni‑modal representation.

\vspace{3pt} \noindent \textbf{Preprocessing strategy.} 
For audio-visual preprocessing, we uniformly partition each video into consecutive temporal intervals. For the $k$‑th interval $[t_k, t_{k+1}]$, we sample a single video frame at the midpoint $t^v_k = (t_k + t_{k+1})/2$, and pair it with the audio segment spanning the entire interval. This midpoint sampling yields a visually centered snapshot within its local acoustic context, providing stable and temporally consistent cross‑modal alignment.
To balance temporal resolution and computational cost, we adopt a dynamic interval length: 
0.5 seconds for videos shorter than 30 seconds and 1 second otherwise.

\vspace{3pt} \noindent \textbf{Omni-modal backbone.}
We adopt Qwen-2.5-Omni-7B-Thinker~\cite{xu2025qwen2} as our backbone. Video frames $\mathcal{V}$ are split into patch grids and encoded via a Vision Transformer (ViT) into visual tokens, while raw audio $\mathcal{A}$ is converted to mel-spectrograms $\mathcal{M}$ and encoded into audio tokens. 
Both visual and audio streams are processed in block-wise chunks ({\em e.g.}, 2s), producing sequences $\{e^v_1, e^v_2, ..., e^v_n\}$ and $\{e^a_1, e^a_2, ..., e^a_n\}$ of equal length, where each chunk contains tokens corresponding to the same temporal interval. Visual, audio, and text tokens $T$ are then interleaved as $[e^v_1, e^a_1, \dots, e^v_n, e^a_n, T]$ and embedded using TMRoPE\cite{xu2025qwen2}, 
which extends rotary position embeddings to jointly encode temporal, height, and width coordinates for cross-modal synchronization. The unified token sequence is fed into an LLM to obtain high-level semantic representations and generate text outputs.

\subsection{Training Strategy}

We train the omni-modal model in an auto-regressive fashion using a two-stage curriculum that gradually strengthens multimodal perception, temporal grounding, and instruction-following. The first stage focuses on omni-modal caption generation to align audio and visual representations and capture fine-grained cross-modal correspondence. 
The second stage performs instruction tuning on diverse video-audio tasks to enhance reasoning, controllability, and robustness to modality-specific perturbations.

\vspace{3pt}\noindent\textbf{Stage I: omni-modal caption generation.}
Using the \textbf{\texttt{AVDC}} dataset, we freeze the LLM backbone and update only the visual and audio encoders to specialize them for omni-modal captioning. 
Each video includes a global caption, and multiple timestamped segment-level audio-visual captions, enabling two complementary training objectives:
(i) \emph{global captioning}, where the model summarizes the entire video, and
(ii) \emph{grounded captioning}, where we prepend the temporal span (e.g., ``From 12.5 to 18.0 seconds:'') to condition the model on a specific interval.
We jointly optimize the model on a mixture of these tasks, encouraging both holistic scene understanding and fine-grained temporal localization. This stage yields the \textit{Ours-Caption} model, which establishes a well-aligned audio-visual representation space under the frozen language prior.

\vspace{3pt}\noindent\textbf{Stage II: omni-modal instruction tuning.}
We then perform instruction tuning on an omni-modal dataset constructed around \textbf{\texttt{AVDC-QA-CoT}}, which provides rich audio-visual question-answer pairs augmented with explicit chain-of-thought rationales. To strengthen disentangled audio reasoning and reduce spurious correlations, we introduce controlled audio-visual misalignment examples by injecting out-of-context sounds, forcing the model to attend to the correct modality when answering. We further augment the corpus with existing video and audio QA benchmarks, and subsample them through a balancing pipeline that accounts for video duration, question type, answer complexity, and modality dependence (vision-only, audio-only, or audio-visual).

The model is trained jointly on this curated mixture using a dual-mode supervision scheme. In \emph{thinking mode}, the model is prompted to generate explicit CoT reasoning enclosed in \texttt{<think>}~\dots~\texttt{</think>} before producing the final answer within \texttt{<answer>}~\dots~\texttt{</answer>}. In \emph{non-thinking mode}, it is supervised to output only the final answer without exposing intermediate reasoning. At inference time, we denote the non-thinking and thinking variants as \textit{Ours-Instruct} and \textit{Ours-Thinking}, respectively. This staged training strategy not only improves raw performance on multimodal QA, but also yields a controllable model that can switch between fast answer generation and interpretable reasoning.

\section{Experiments}
We empirically evaluate our trained omni-modal models across a suite of audio–visual understanding tasks. Our evaluation covers
(i) audio–visual captioning, to assess coherent cross-modal perception;
(ii) omni-modal question answering, to measure multimodal comprehension and reasoning; and
(iii) audio-only question answering, to specifically probe audio-centric understanding and the model’s ability to disentangle auditory cues from visual context.

\subsection{Performance on Omni-Modal Captioning}

\noindent \textbf{Benchmarks.} We evaluate omni-modal captioning on three captioning benchmarks: 
our \textbf{\texttt{AVDC-test}} split, and two existing benchmarks, Video-SALMONN-2 and UGC-VideoCap. \textbf{\texttt{AVDC-test}} consists of real-world videos randomly sampled from \textbf{\texttt{AVDC}}, covering diverse and often cluttered acoustic scenes. To build its annotations, we use GPT-4o to extract both visible (with clear visual evidence) and invisible (without visual evidence) sound events from the sampled audio-visual captions, followed by manual verification to ensure accuracy and consistency.

\textbf{\texttt{AVDC-test}} evaluates whether generated captions can recover both visible and invisible sound events, thus probing auditory grounding with and without visual support. Video-SALMONN-2 focuses on detailed captioning and measures whether model outputs include the ground-truth audio and visual events. UGC-VideoCap targets user-generated content and assesses overall caption quality along audio, visual, and detail dimensions relative to human-written references.

\vspace{3pt} \noindent \textbf{Metrics.} On \textbf{\texttt{AVDC-test}}, we report the missing rates of reference visible and invisible sound events, denoted Miss$_{\text{vis}}$ and Miss$_{\text{inv}}$, as well as the overall missing rate Miss. On the Video-SALMONN-2 test set, we follow the original protocol and compute the missing rate (Miss\%) and hallucination rate (Hall\%) of audio and visual events with respect to ground-truth annotations, using GPT-3.5 for automatic judgment. For UGC-VideoCap, we adopt GPT-4o to score caption quality in terms of audio, visual, and detail aspects on a 1–5 scale and convert these scores to percentages. All evaluations on Video-SALMONN-2 and UGC-VideoCap strictly follow the setups defined in their respective papers.

\vspace{3pt} \noindent \textbf{Results.} 
Table~\ref{tab: caption benckmarks} summarizes the quantitative results and leads to three main observations:
\textbf{i) proprietary models remain very strong.} Proprietary omni-modal systems consistently perform well across all benchmarks. They capture a broad range of audio and visual events and generate fluent, semantically rich descriptions, which translates into low missing rates and high caption quality scores.
\textbf{ii) open-source baselines lag behind, especially on invisible audio.}
Existing open-source models struggle to detect and localize sound events, particularly those without visual evidence. On \textbf{\texttt{AVDC-test}}, nearly 90\% of invisible sound events are missed. Moreover, they exhibit a pronounced trade-off between coverage and precision: models that aggressively reduce missing rates tend to hallucinate many non-existent events, while models that suppress hallucinations typically fail to cover key content, especially in the audio stream.
\textbf{iii) our model improves both audio coverage and balance.}
Our omni-modal model achieves the lowest missing rates for both visible and invisible sound events on \textbf{\texttt{AVDC-test}}, reducing the overall miss rate by more than 10 points, compared with the Qwen2.5-Omni baseline after our continued training. 
On UGC-VideoCap, it delivers more balanced multimodal perception with clear gains in the audio dimension, indicating stronger auditory grounding in unconstrained user-generated videos. On Video-SALMONN-2, our model simultaneously decreases both missing and hallucination rates, resulting in the best overall trade-off among all systems and demonstrating that our training strategy improves not only recall of true events but also robustness to spurious cross-modal correlations, leading to more reliable and faithful audio-visual understanding.

\begin{table*}[!htb]
\setlength{\tabcolsep}{7pt} 
\centering
\Large 
\caption{Performance comparison of different models on captioning benchmarks. In our AVDC-test, \textit{Miss$_{\text{vis}}$} and \textit{Miss$_{\text{inv}}$} denote the missing rates of visible and invisible sound events relative to all reference events. For the video-SALMONN-2 testset, \textit{Miss} and \textit{Hall.} represent the percentages of missed and hallucinated events relative to the annotated ground truth. In UGC-VideoCap, \textit{Audio}, \textit{Visual}, and \textit{Detail} indicate quality scores assessed by GPT-4o with reference to the provided captions. \textit{Ours-Caption} and \textit{Ours-Instruct} denote the models fine-tuned on captions in Stage 1 and further instruction-tuned in Stage 2, respectively. The best-performing models in each category are highlighted in \textbf{bold}, and the second-best ones are \underline{underlined}.}
\vspace{-0.3cm}
\resizebox{\linewidth}{!}
{
\begin{tabular}{ll|ccc|cccc|ccc}
\toprule
\multirow{2}{*}{Models} & \multirow{2}{*}{\shortstack{LLM\\ Size}} 
 & \multicolumn{3}{c|}{AVDC-test}  & \multicolumn{4}{c|}{UGC-VideoCap}    & \multicolumn{3}{c}{video-SALMONN-2 testset}                                                                                      \\ \cline{3-12} 
& 
& Miss$_{\text{vis}}$$\downarrow$ 
& Miss$_{\text{inv}}$$\downarrow$ 
& Miss$\downarrow$ 
 & Audio$\uparrow$ & Visual$\uparrow$ & Detail$\uparrow$ & Avg.$\uparrow$ 
 & Miss$\downarrow$        & Hall.$\downarrow$        & Total$\downarrow$  \\ 
\midrule
\multicolumn{12}{c}{\cellcolor{blue!10}Proprietary Models}  \\ 
\midrule
Gemini-2.5-Flash & -  & 31.8 & 50.3 & 39.1  & 74.2   & 78.8    & 77.2    & 76.7  & 19.3  & 13.9   & 33.3 \\
Gemini 2.5 Pro   & -  & 26.7 & 46.7 & 34.8  & 70.8   & 75.8    & 74.8    & 73.8        & 18.1  & 13.3   & 31.3\\
\midrule
\multicolumn{12}{c}{\cellcolor{orange!10}Open-Source Models}  \\ 
\midrule 
VideoLLaMA 2~\cite{cheng2024videollama}     & 7B  & 75.2 & 95.1 & 83.0 & -      & -       & -  & -   & 56.8 & \textbf{8.9}  & 65.7  \\
MiniCPM-o-2.6~\cite{team2025minicpm}    & 7B  & 70.2 & 90.4 & 77.5   & 38.6   & 68.5    & 57.7    & 54.9 & 42.2  & 14.3   & 56.5  \\
HumanOmniV2~\cite{yang2025humanomniv2}     & 7B  & 68.6 & 90.9 & 77.1  & 45.6   & 66.3    & 59.5    & 57.1  & 49.2  & 12.3   & 61.6 \\
video-SALMONN 2~\cite{tang2025video}  & 7B & \underline{60.6} & \underline{87.3} & \underline{70.9} & 61.8   & \textbf{71.4}    & \textbf{68.5}    & \underline{67.2}   & \textbf{21.2}  & 17.6   & \underline{38.8}  \\ 
\midrule
\rowcolor{lightgray} Qwen2.5-Omni~\cite{xu2025qwen2}     & 7B  & 60.5 & 90.7 & 72.1  & 46.9   & 66.1    & 60      & 57.7  & 41.7  & 15.4   & 57.1  \\			
\rowcolor{lightgray} Ours-Caption  & 7B  & 50.3 & 79.2 & 61.9   & 62.7	& 68.8	& 62.4	& 64.6  & 27.8  & 11.8   & 39.6 \\ 
\rowcolor{lightgray} Ours-Instruct    & 7B  & \textbf{48.0} &	\textbf{77.5} & \textbf{59.5}
 & \textbf{65.4}   & \underline{71.2}      & \underline{67.7}    & \textbf{68.1} & \underline{26.4}    & \underline{11.2}    & \textbf{37.6}  \\ 
\bottomrule
\end{tabular}
}
\label{tab: caption benckmarks}
\end{table*}

\subsection{Results on Video QA Benchmarks}\label{sec: omni benchmark}

\noindent \textbf{Benchmarks.} 
We evaluate our models on four omni-modal video understanding benchmarks: WorldSense~\cite{hong2025worldsense}, Daily-Omni~\cite{zhou2025daily}, Omnibench~\cite{li2024omnibench}, and ACVUBench~\cite{yang2025acvubench}, and two visual-centric benchmarks: Video-MME~\cite{fu2025video} and MVBench~\cite{li2024mvbench}. 
OmniBench contains 1,142 samples that jointly evaluate visual, audio, and textual understanding within a unified assessment suite.
WorldSense and Daily-Omni contain 3,172 and 1,197 multiple-choice QA pairs, respectively, covering diverse real-world scenarios and emphasizing coordinated use of omni-modal information and audio–visual alignment. 
ACVUBench comprises over 10k video-based QA pairs designed to probe detailed audio understanding and audio–visual interactions, including subtle sound events, temporal relations, and cross-modal consistency. Video-MME and MVBench are widely used video benchmarks but mainly assess visual understanding rather than audio-visual reasoning.

\vspace{3pt} \noindent \textbf{Metrics.} 
We report multiple-choice accuracy on all benchmarks. Each model is evaluated in two inference modes: the \textbf{standard inference mode}, in which the model directly outputs its choice, and the \textbf{thinking mode}, the model first generates an explicit reasoning trace and then selects its final answer based on this reasoning. This allows us to assess both raw predictive performance and the effect of explicit chain-of-thought reasoning on omni-modal QA.

\vspace{3pt} \noindent \textbf{Results.} 
Table~\ref{tab: omni benchmark} summarizes the performance across benchmarks and leads to three main observations:
\textbf{i) proprietary models remain the strongest overall.}
Commercial omni-modal models achieve the highest average accuracy, consistent with their large-scale pretraining and heavily optimized multimodal integration pipelines.
\textbf{ii) our training strategy yields consistent gains over the base model.}
Starting from Qwen2.5-Omni, our instruction-tuned model obtains at least 3 points of accuracy improvement on all four benchmarks. These gains indicate that our \textbf{\texttt{AVDC}} and \textbf{\texttt{AVDC-QA-CoT}} supervision—especially the explicit audio-visual grounding and CoT-based reasoning signals—substantially enhance general omni-modal understanding. We also evaluate our model on vision-centric benchmarks, Video-MME and MVBench, where it achieves near-identical accuracy to the original performance. This result indicates that the improvements in audio-visual understanding do not compromise visual-only capabilities.
\textbf{iii) thinking mode offers additional benefits, rivaling strong proprietary systems.}
When evaluated in thinking mode, our model matches or exceeds prior open-source approaches on all benchmarks and reaches parity with Gemini-2.5-Flash on WorldSense. This suggests that combining rich audio-visual cues with explicit reasoning not only improves accuracy in complex, fine-grained scenarios, but also narrows the gap between open-source omni-modal models and state-of-the-art proprietary systems.

\begin{table}[t]
\vspace{-0.2cm}
\setlength{\tabcolsep}{7pt} 
\centering
\large
\caption{Performance comparison of different models on omni-modal benchmarks. All benchmarks are evaluated based on the accuracy of multiple-choice problems. \textit{Ours-Instruct} and \textit{Ours-Thinking} infer in non-thinking and thinking modes, respectively. The best-performing models in each category are highlighted in \textbf{bold}, and the second-best ones are \underline{underlined}. Scores obtained under the \textit{thinking mode} are shown in \blue{blue}.}
\vspace{-0.3cm}
\scalebox{0.58}{
\begin{tabular}{ll|c|c|c|c|c|c}
\toprule
\multirow{2}{*}{Models} & \multirow{2}{*}{\shortstack{LLM\\ Size}}  &
\multicolumn{4}{c|}{Omni-Modal}  &
\multicolumn{2}{c}{Visual-Centric} \\
\cline{3-8} 
& & WorldSense & Daily-Omni & Omnibench & AVUT & Video-MME & MVBench \\ 
\midrule
\multicolumn{8}{c}{\cellcolor{blue!10}Proprietary Models}  \\ 
\midrule
Gemini-2.5-Flash & -  & \textcolor{blue}{52.3}  & 72.7 &- &- & 81.5 &- \\
Gemini 2.5 Pro &-  & \textcolor{blue}{65.1} &- &- &- & 86.9 &- \\
\midrule
\multicolumn{8}{c}{\cellcolor{orange!10}Open-Source Models}  \\ 
\midrule
VideoLLaMA 2~\cite{cheng2024videollama}     & 7B & 25.4 & 35.17  & - & 40.56 & 47.9 & 54.6   \\
MiniCPM-o-2.6~\cite{team2025minicpm}    & 7B  & - &53.21  & 40.5 &- & 63.9 &- \\
HumanOmniV2~\cite{yang2025humanomniv2}     & 7B & 47.1 & 58.5 & - & - &- &- \\

video-SALMONN 2~\cite{tang2025video}  & 7B  & \underline{48.6} & \underline{66.2} & - & \underline{65.6} & \textbf{67.4} &- \\ 
\midrule
\rowcolor{lightgray} Qwen2.5-Omni~\cite{xu2025qwen2}  & 7B  & 45.4 & 47.5 & \underline{56.1} & 61.6 & 64.3 & \underline{70.3} \\
\rowcolor{lightgray} Ours-Instruct    & 7B & 50.8 & 65.3  & 58.0  & 67.1 & 65.6 & 70.1 \\ 
\rowcolor{lightgray} Ours-Thinking    & 7B & \textbf{\textcolor{blue}{52.4}} & \textbf{\textcolor{blue}{67.6}} & \textbf{\textcolor{blue}{59.3}} & \textbf{\textcolor{blue}{68.0}} & \underline{\textcolor{blue}{65.8}} & \textbf{\textcolor{blue}{70.4}} \\ 

\bottomrule
\end{tabular}
}
\label{tab: omni benchmark}
\vspace{-0.5cm}
\end{table}

\subsection{Results on Audio QA Benchmarks}
\noindent \textbf{Benchmarks.} 
We further assess audio-centric understanding on two audio-only QA benchmarks: 
MMAU (Test-mini)~\cite{sakshi2024mmau} and MMAR~\cite{ma2025mmar}. 
MMAU (Test-mini) contains 10k QA pairs covering a broad range of auditory phenomena, including speech, environmental sounds, and music. MMAR also consists of 10k QA pairs, but emphasizes mixed auditory conditions with varying combinations of sound effects, music, and speech. Both benchmarks are evaluated using the same multiple-choice accuracy protocol and inference modes described in Section~\ref{sec: omni benchmark}.

\vspace{3pt} \noindent \textbf{Results.} 
Table~\ref{tab: audio benchmark} reports the quantitative results and yields two main observations:
\textbf{i) gains over the omni-modal backbone.} 
Our model consistently outperforms its backbone, Qwen2.5-Omni, across both MMAU and MMAR. This indicates that our curated audio-visual captioning and QA data—in particular, the explicit modeling of invisible sounds and controlled audio–visual misalignment—effectively enrich the model’s auditory representations and robustness to diverse acoustic conditions.
\textbf{ii) competitive with specialized audio-language systems.}
Overall, our model achieves competitive or superior performance compared with strong audio-language baselines, surpassing most open-source systems on both benchmarks. These results suggest that our multimodal training strategy not only avoids degrading audio understanding through overreliance on visual information, but in many cases improves pure audio comprehension, even when no visual input is available at inference time.

\begin{table}[t]
\vspace{-0.4cm}
\setlength{\tabcolsep}{7pt} 
\centering
\large
\caption{Accuracy performance comparison of different models on existing audio understanding benchmarks. \textit{Ours-Instruct} and \textit{Ours-Thinking} infer in non-thinking and thinking modes, respectively. The best-performing models in each category are highlighted in \textbf{bold}, while the second-best results are \underline{underlined}. Scores obtained under the \textit{thinking mode} are shown in \blue{blue}.}
\vspace{-0.3cm}
\scalebox{0.6}{
\begin{tabular}{p{13.0em}C{5.0em}C{5.0em}|C{5.0em}C{5.0em}}

\toprule

Models & Modality & LLM Size & MMAR & MMAU \\
\midrule

\multicolumn{5}{c}{\cellcolor{blue!10}Proprietary Models}  \\ 
\midrule
Gemini-2.5-pro       & A+V      & -& -    & 71.6  \\
Gemini-2.5-Flash     & A+V      & - & -    & 71.8  \\

GPT-4o Audio         & A        & -& 63.5 & 62.5  \\ \midrule

\multicolumn{5}{c}{\cellcolor{orange!10}Open-Source Models}  \\
\midrule

SALMONN~\cite{tang2023salmonn}              & A        & 13B    & 33.2 & 57.1 \\
Qwen2-Audio-Instruct~\cite{chu2024qwen2} & A        & 7B         & 30   & 59.6  \\
Kimi-Audio~\cite{ding2025kimi}           & A        & 8.2B   & -    & 68.2  \\
Audio Flamingo 3~\cite{goel2025audio}     & A        & 8.2B    & -    & \textbf{73.3}  \\

Baichuan-Omni-1.5~\cite{li2025baichuan}    & A+V      & 11B      & 40.7 & -     \\ 
\midrule
\rowcolor{lightgray} Qwen2.5-Omni~\cite{xu2025qwen2}         & A+V      & 7B        & \underline{53.8} & 71.5  \\
\rowcolor{lightgray} Ours-Instruct        & A+V      & 7B        & 54.3 & 72.1  \\ 
\rowcolor{lightgray} Ours-Thinking        & A+V      & 7B         & \textbf{\blue{55.4}} & \underline{\blue{72.9}}
\\ 
\bottomrule
\end{tabular}
}
\label{tab: audio benchmark}
\vspace{-0.4cm}
\end{table}

\section{Related Works}
\label{sec:related_works}
\noindent \textbf{Audio-visual dataset.}
Recent progress in video understanding has driven the creation of large-scale audio-visual caption datasets to enhance omni-modal perception. VAST-27M~\cite{chen2023vast} trains captioners and applies LLM to generate omni-modal descriptions, while VALOR-1M~\cite{chen2023valor} provides manually annotated captions. AVCaps~\cite{sudarsanam2025avcaps} employs LLM to produce captions by integrating crowdsourced audio, visual, and audio-visual annotations. Other datasets emphasize specific aspects, such as music understanding (MMtrail~\cite{chi2024mmtrail}, HarmonySet~\cite{zhou2025harmonyset}) and speech understanding (YODAS~\cite{li2023yodas}, LiveCC~\cite{chen2025livecc}).
In parallel, audio-visual question answering (AVQA) datasets have been developed to advance omni-modal reasoning, including AVQA~\cite{yang2022avqa} for real-world scenes, MUSIC-AVQA~\cite{li2022learning} for music-based spatiotemporal reasoning, and AV-UIE~\cite{du2025crab} for unified instruction tuning across spatial and temporal tasks.

However, recent audio-visual caption datasets provide concise annotations lacking fine-grained temporal and semantic details~\cite{chen2023vast,chen2023valor}, rely heavily on costly manual labeling~\cite{chen2023valor,sudarsanam2025avcaps,wu2025ugc}. Furthermore, most works mainly focus on naturally aligned events~\cite{geng2025longvale,cheng2024avset}, overlooking decoupled cases where sounds occur without visible sources or visual cues induce auditory illusions.

\vspace{3pt}\noindent \textbf{Omni-modal models.}
Omni-modal models aim to perceive, understand, and generate content across multiple modalities (e.g., text, image, video, audio), supporting flexible multimodal inputs and outputs. 
The Google Gemini~\cite{team2023gemini} model performs video understanding as a native omni-modal LLM, integrating tokens from text, audio, and visual streams.
Video-LLaMA~\cite{zhang2023video} and VideoLLaMA2~\cite{cheng2024videollama} achieve video joint understanding by concatenating audio and visual tokens. Recent omni-modal models such as Qwen2.5-Omni~\cite{xu2025qwen2}, Qwen3-Omni~\cite{xu2025qwen3}, and Baichuan-Omni-1.5~\cite{li2025baichuan} can process modalities simultaneously and generate text or speech in a streaming manner. Ola~\cite{liu2025ola} improves omni-modal alignment through a progressive strategy, while Video-SALMONN 2~\cite{tang2025video} advances omni-modal learning via caption enhancement. AVoCaDO~\cite{orchestrationavocado} and Omni-Captioner~\cite{ma2025omni} are mainly developed for captioning and do not generalize to instruction-following tasks.

Despite recent progress, key challenges remain. Omni-modal models still struggle with fine-grained temporal alignment in video understanding~\cite{zhou2025daily, geng2025longvale}, especially when dealing with rapidly changing scenes and asynchronous cross-modal cues. While recent models~\cite{xu2025qwen2, xu2025qwen3, li2025baichuan} have improved speech and music comprehension, their understanding of real-world mixed sounds-often composed of overlapping, noisy, and context-dependent audio events-remains limited. Moreover, most existing systems primarily operate on short video inputs\cite{zou2024seconds, madan2024foundation}, highlighting the need to extend their capability to understand longer videos.

\section{Conclusion}

This work advances omni-modal understanding by explicitly strengthening auditory perception within a unified video–audio–language framework. 
To start, we introduce \textbf{\texttt{AVDC}}, a large-scale caption dataset that both decouples and integrates audio and visual semantics via an LLM-driven pipeline, and \textbf{\texttt{AVDC-QA-CoT}}, an instruction-tuning dataset with chain-of-thought rationales that supports detailed and interpretable audio-visual reasoning.
Leveraging these resources, we propose a two-stage training strategy: omni-modal caption generation to align and enrich modality-specific representations, followed by instruction tuning to improve cross-modal reasoning and controllable inference. Extensive experiments on captioning, audio-only QA, and omni-modal QA benchmarks demonstrate consistent performance gains over strong baselines, particularly in recognizing invisible sounds and resolving audio-visual interactions. These results highlight the importance of dedicated auditory supervision and structured reasoning for building more reliable omni-modal foundation models.


\clearpage  


%
%
\bibliographystyle{splncs04}
\bibliography{main}

\clearpage
\setcounter{page}{1}



\section{Supplementary}

\subsection{AVDC Dataset Statistics}
In this section, we conduct a systematic analysis of the proposed AVDC dataset from several complementary perspectives, including sound event distribution, video duration, and caption length. And we find: (i)AVDC covers a broad and fine-grained set of acoustic categories, explicitly distinguishing between visible and invisible (off-screen or occluded) sound events; (ii)AVDC consists of long-duration videos paired with detailed, linguistically rich captions, thereby providing ample temporal and semantic context for modeling complex audio-visual interactions.

\begin{figure}[htbp]
    \centering
    \includegraphics[width=0.8\linewidth]{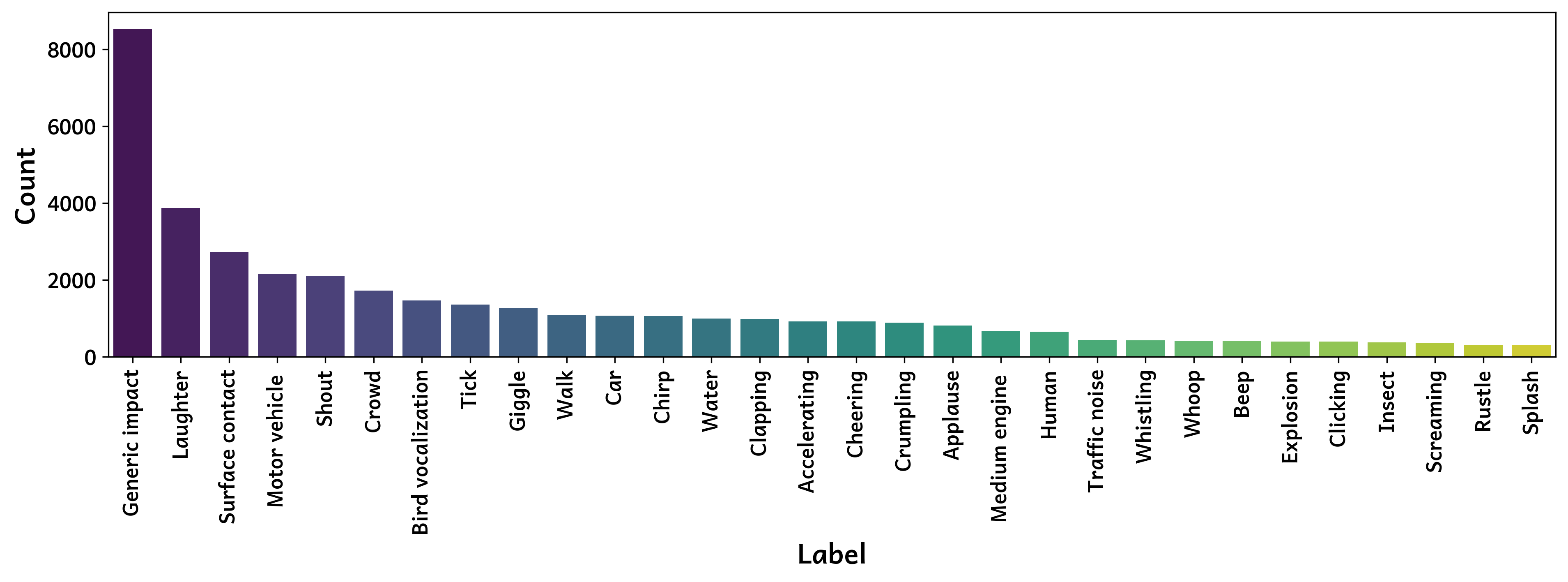}
    \vspace{-1.0em}
    \caption{Top-30 sound event labels}
    \label{fig:sed_counts}
    \vspace{-0.6em}
\end{figure}

\vspace{2pt}\noindent\textbf{Sound event.}
We present the distribution of the top 30 sound event labels in Figure~\ref{fig:sed_counts}. We use ATST-F~\cite{schmid2025effective} to detect all sound events in videos, leveraging the Audioset~\cite{gemmeke2017audio} label set.
Generic impact sounds are the most frequent, containing a broad coverage of object collisions, drops, and hits. They are followed by Laughter, Surface Contact events (e.g., Scrape), and Shout. Overall, the dataset spans a wide and diverse range of acoustic categories.

\begin{figure}[htbp]
    \centering
    \begin{minipage}{0.49\linewidth}
        \begin{minipage}[b]{0.46\textwidth}
            \centering
            \includegraphics[width=\linewidth]{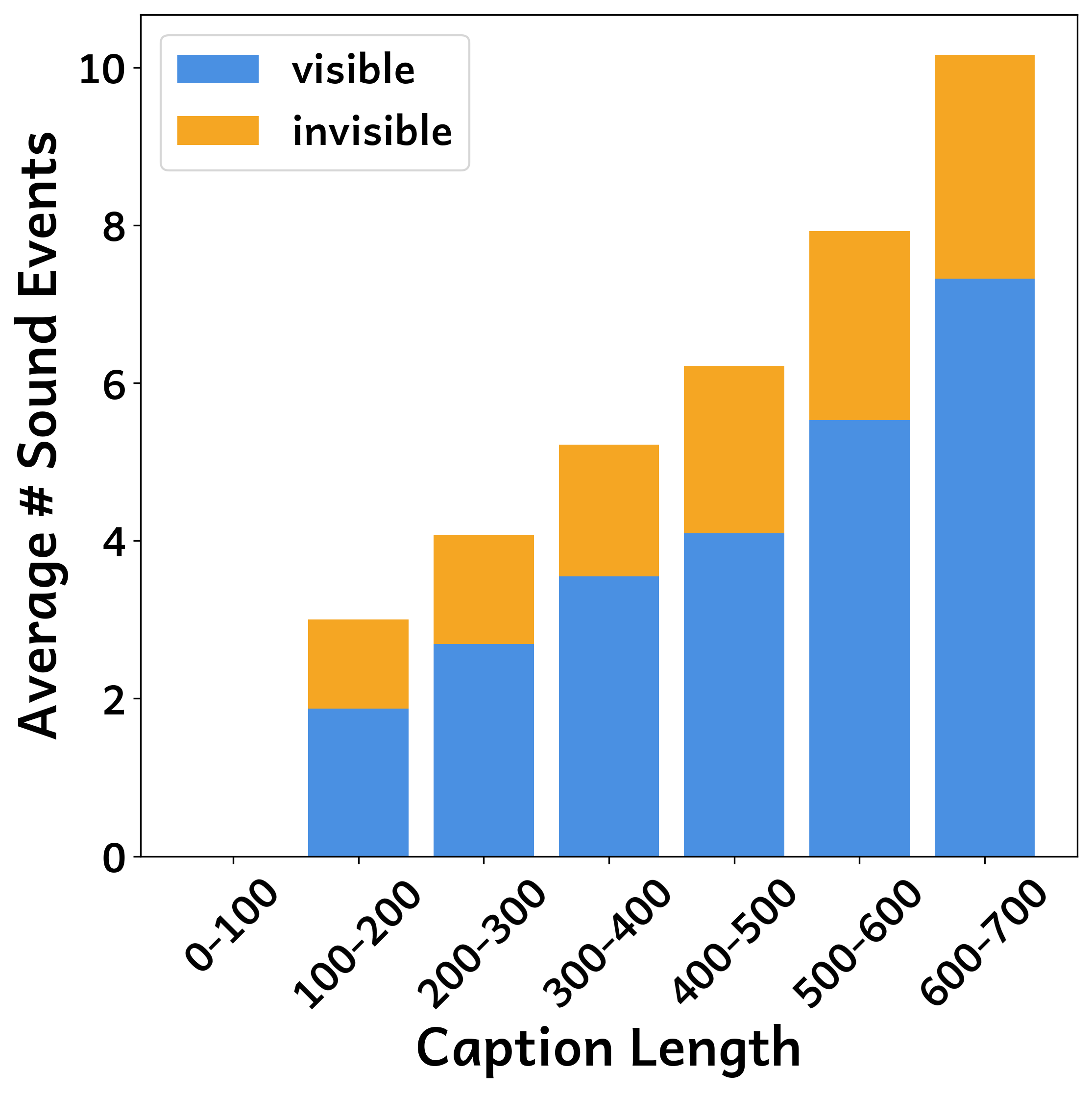}
        \end{minipage}
        \hspace{-0.2em}
        \begin{minipage}[b]{0.46\textwidth}
            \centering
            \includegraphics[width=\linewidth]{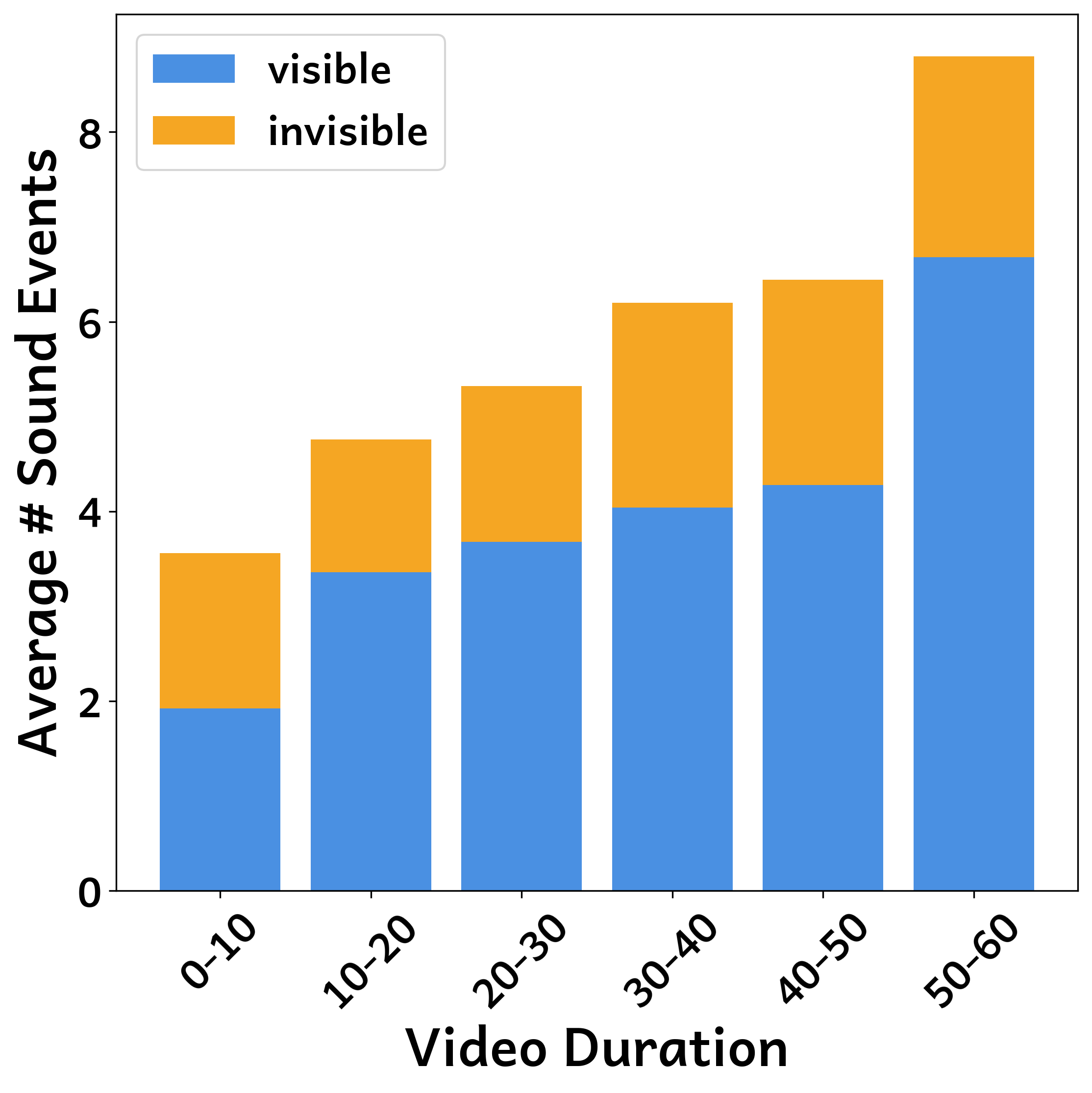}
        \end{minipage}
    \vspace{-0.5em}
    \caption{Statistics of visible and invisible sound event labels}
    \label{fig:seds}
    \end{minipage}
    \begin{minipage}{0.49\linewidth}
        \hspace{-0.2em}
        \begin{minipage}[b]{0.46\textwidth}
            \centering
            \includegraphics[width=\linewidth]{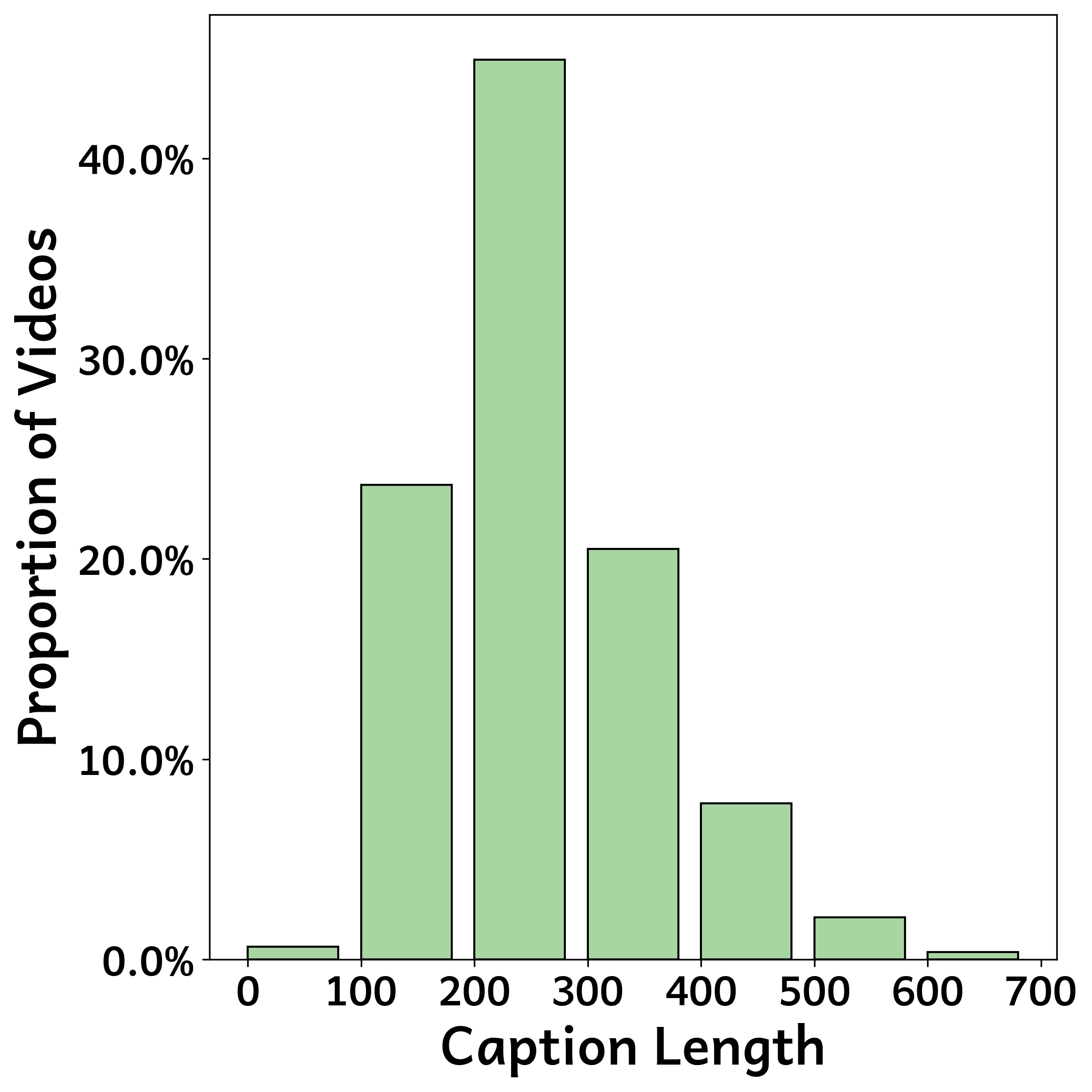}
        \end{minipage}
        \hspace{-0.2em}
        \begin{minipage}[b]{0.46\textwidth}
            \centering
            \includegraphics[width=\linewidth]{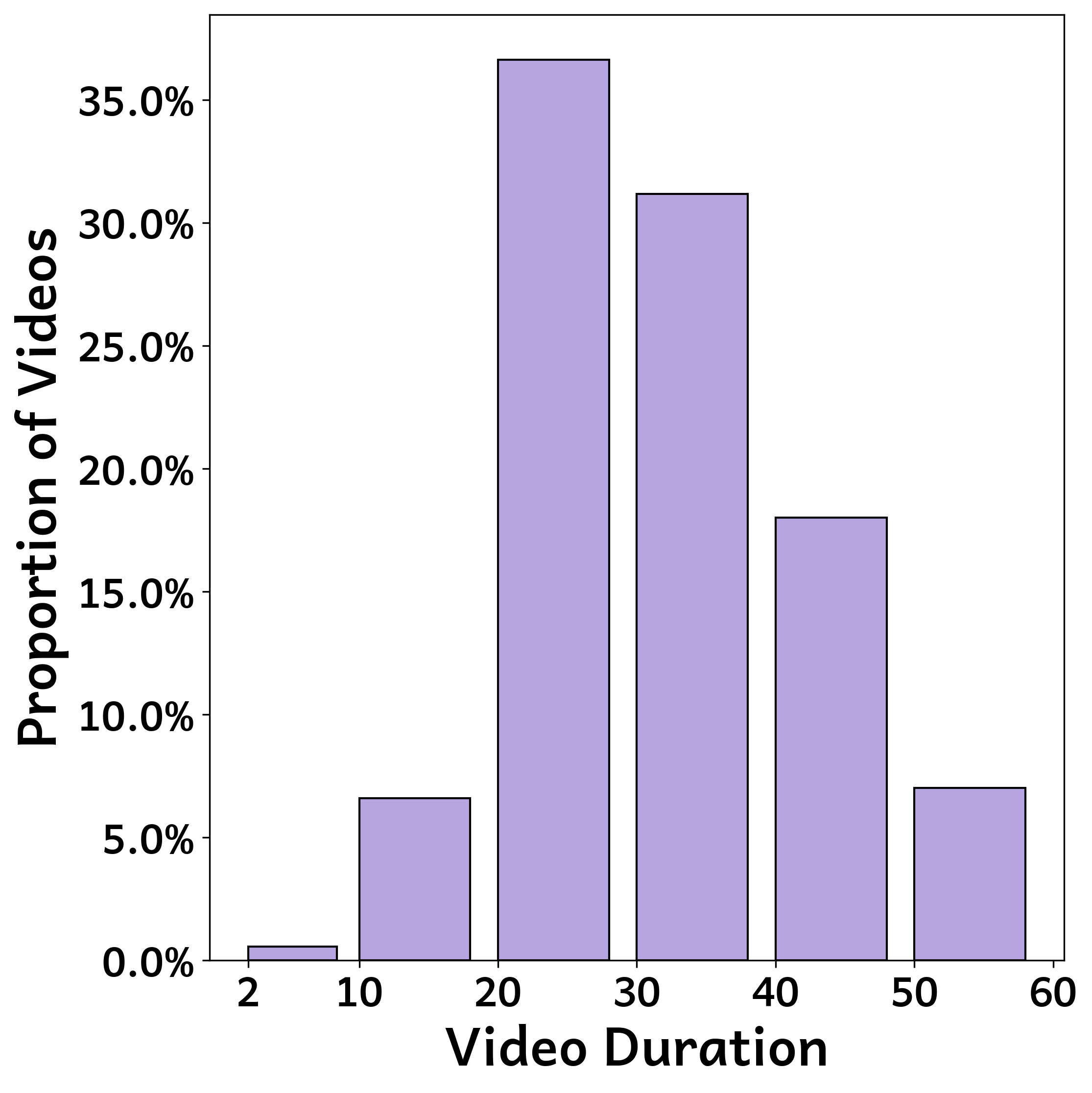}
        \end{minipage}
    \vspace{-0.5em}
    \caption{Distribution of Caption Lengths and Video Durations}
    \label{fig:lens}
    \end{minipage}
\vspace{-0.5em}
\end{figure}

We use GPT-4o to detect and categorize sound events as visible or invisible based on audio-visual captions. Figure~\ref{fig:seds} shows their distribution across caption lengths and video durations, and we draw three key observations: (i) Both the average number of sound events and the proportion of invisible sounds grow with caption length, suggesting that longer captions capture richer audio content and more non-visual cues. (ii) The number of sound events increases with video duration, indicating more acoustic information in longer videos. (iii) The proportion of invisible sounds remains stable across video durations, demonstrating a balanced mix of visible and invisible events in the dataset.

\vspace{2pt}\noindent\textbf{Caption length and video length.}
We report statistics on video duration and caption characteristics to quantify the temporal and linguistic complexity of AVDC in Figure~\ref{fig:lens}. We observe that (i)~about 90\% of captions contain 100-400 words, with a maximum of 695, forming an even distribution without extreme outliers; and (ii)~roughly 90\% of videos last 20-50 seconds, with a maximum of 60 seconds. Overall, we find that AVDC offers a well-balanced mix of textual richness and temporal coverage, with captions that are semantically informative and videos long enough to capture meaningful event dynamics.




\subsection{Experimental Details}
In this section, we present the hyperparameters for the two-stage training and the composition of the instruction-tuning data.

\subsubsection{Training Details}
Our training strategy adopts a two-stage paradigm. In Stage~I, we freeze the language LLM and train only the vision and audio encoders on the omni-caption generation task to strengthen audio-visual perception. In Stage~II, we unfreeze all parameters for instruction tuning, leveraging a broad range of VQA, AQA, and AVQA data to improve instruction following and reasoning capabilities. Table~\ref{tab:training} summarizes the hyperparameters used in both stages.

\vspace{-0.3em}
\begin{table}[htbp]
\centering
\caption{Hyperparameters for two-stage training.}
\vspace{-0.6em}
\scalebox{1.0}{
\begin{tabular}{c|cc}
\hline
                      & Stage~I       & Stage~II      \\ \hline
Used GPUs             & \multicolumn{2}{c}{4 x H200s} \\
Batch Size per GPU    & 2             & 1             \\
Gradient Accumulation & 4             & 4             \\
Peak Learning Rate    & 1e-5          & 5e-6          \\
Updates               & 2000          & 7906          \\
Training Time         & 30h           & 114h          \\ \hline
\end{tabular}
}
\vspace{-0.8em}
\label{tab:training}
\end{table}

\subsubsection{Composition of the Instruction-Tuning Data}
\begin{figure*}[!t]
    \centering
    \includegraphics[width=0.85\linewidth]{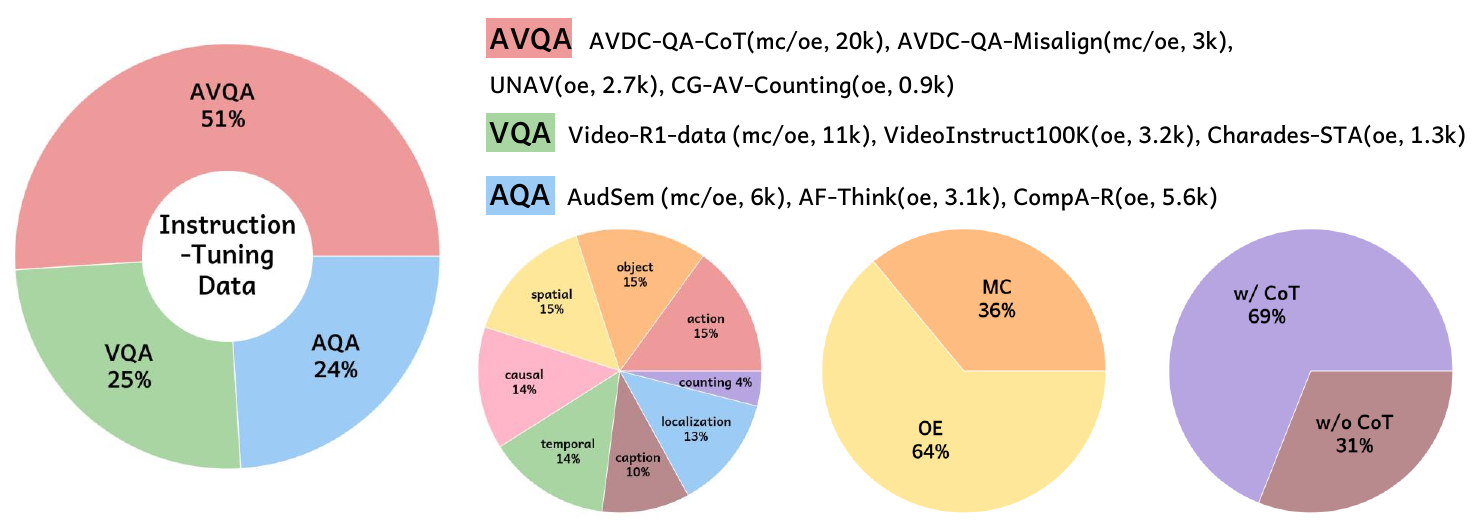}
    \vspace{-0.5em}
    \caption{Data statistics of the instruction-tuning data. It covers a broad range of question types, including those with or without chain-of-thought reasoning, as well as multiple-choice and open-ended questions. Abbreviations: AVQA(audio-visual QA), VQA(video QA), AQA(audio QA), MC (multiple-choice), OE(open-ended).}
    \label{fig:instruction_data}
    \vspace{-0.5em}
\end{figure*}

Figure~\ref{fig:instruction_data} summarizes our instruction-tuning corpus, which consists of 51\% AVQA, 25\% VQA, and 24\% AQA samples. The AVQA portion combines our AVDC-QA with UNAV~\cite{geng2023dense} and CG-AV-Counting~\cite{lu2025av}; the VQA portion includes Video-R1-data~\cite{feng2025video}, VideoInstruct100K~\cite{maaz2024video}, and Charades-STA~\cite{gao2017tall}; and the AQA portion incorporates AudSem~\cite{wijngaard2025audsemthinker}, AF-Think~\cite{goel2025audio}, and CompA-R~\cite{ghosh2024gama}. We balance question types across action, object, spatial, causal, temporal, and localization categories (13-15\%), with counting and caption questions at 4\% and 10\%, respectively. We also find that 64\% of samples are open-ended, 36\% are multiple-choice, and 69\% include chain-of-thought rationales. Overall, we provide a diverse, balanced, and reasoning-rich corpus for instruction tuning.

\vspace{-0.3em}
\begin{table}[htbp]
\centering
\caption{Ablation study on training stages and instruction-tuning datasets. V\&A refers to the conbined VQA and AQA datasets.}
\vspace{-0.5em}
\scalebox{0.95}{
\begin{tabular}{ccc|cc|cccc|cc}
\hline
&\multicolumn{2}{c|}{Stage} &\multicolumn{2}{c|}{Modules} & \multicolumn{4}{c|}{Datasets} & \multirow{2}{*}{WorldSense} & \multirow{2}{*}{Daily-Omni} \\ \cline{1-9}
&I       & II  & \begin{tabular}[c]{@{}c@{}}Audio\\ Encoder\end{tabular}  & LLM    & AVDC  & \begin{tabular}[c]{@{}c@{}}AVQA-QA-\\ Aligned\end{tabular} & \begin{tabular}[c]{@{}c@{}}AVQA-QA-\\ Misaligned\end{tabular} & V\&A &                             &                             \\ \hline
1&-             & -      & -     & -        & -     & -    &-   & -          & 45.4                        & 47.5                        \\
2&\cmark  &   &\cmark     &     & \cmark     &         &      &     & 46.8                     & 51.0                          \\
3&  & \cmark   &    & \cmark     &   &  \cmark  & \cmark      &     & 47.3      & 55.7                      \\
4& & \cmark      & \cmark  & \cmark        &    & \cmark  & \cmark       &            & 49.6                        & 61.5                        \\
5& & \cmark    & \cmark  & \cmark         &    &      &   & \cmark          & 45.0                        & 47.6                        \\
6& & \cmark    & \cmark  & \cmark         &     & \cmark    & \cmark    & \cmark          & 49.3                        & 61.8                        \\
7& \cmark & \cmark     & \cmark  & \cmark        & \cmark     &    &     & \cmark          & 48.6                        & 62.4                        \\
8&\cmark  & \cmark   & \cmark  & \cmark    & \cmark     & \cmark  &     &            & 49.8                      & 63.4                       \\
9&\cmark   & \cmark   & \cmark  & \cmark  & \cmark     & \cmark  & \cmark     &            & 50.3                        & 64.8                        \\
10&\cmark   & \cmark  & \cmark  & \cmark  & \cmark     & \cmark  & \cmark     & \cmark          & 50.8                        & 65.3                        \\ \hline
\end{tabular}
}
\vspace{-0.8em}
\label{tab:ablation}
\end{table}

\subsection{Ablation Studies}
To evaluate the effectiveness of AVDC and AVDC-QA, we ablate the training stages and the datasets used for Stage~II instruction tuning, and assess performance on WorldSense~\cite{hong2025worldsense} and Daily-Omni~\cite{zhou2025daily}. From Table~\ref{tab:ablation}, we make three observations: 
(i)~Rows 1-3,9 demonstrate that fine-tuning the LLM with frozen encoders yields only limited gains (e.g., 1.9\% on WorldSense and 8.2\% on Daily-Omni) and consistently underperforms full parameter fine-tuning and our two-stage approach, confirming the necessity of joint encoder-LLM optimization.
(ii)~Rows 4-6 show that Stage~II tuning on generic VQA/AQA yields only minor gains, whereas using AVDC-QA produces substantially larger improvements, highlighting the value of stronger audio-visual correspondence. 
(iii)~Comparing the last three rows with the earlier ones, we find that adding Stage~I AVDC training significantly boosts both benchmarks, demonstrating the importance of large-scale omni-modal caption supervision for robust representations. 
(iiii)~Within the last three rows, combining AVDC-QA with VQA/AQA in Stage~II achieves the best results, showing that mixed instruction data provides more diverse task signals and richer reasoning cues. Overall, these findings validate our two-stage paradigm and the important roles of AVDC, AVDC-QA, and VQA/AQA supervision.




\subsection{Prompts and Data Visualization}
In this section, we present the prompts used in our automatic pipeline for audio-visual caption generation, covering segment-level/video-level captions generation and verification, and caption-based QA generation. 

\begin{casestudy}{Segment-level Audio Caption Generation Prompt}
I will give you some information from a video and its audio, audio is extracted from the video.

\medskip

For the previous segment:\\
The audio caption is:\\
\{\textit{audio caption in previous segment}\}

\medskip

For the current segment from \{\textit{timestamp[0]}\} to \{\textit{timestamp[1]}\} seconds:\\
The video caption is:\\
\{\textit{video caption in current segment}\}

\medskip

The music information for audio is:\\
\{\textit{music caption in current segment}\}

\medskip

The speech information for audio is:\\
\{\textit{speech caption in current segment}\}

\medskip

The audio information for reference is: \\
\{\textit{audio caption in current segment}\}

The verification for the above audio information is: \\
\{\textit{audio caption verification in current segment}\}

\medskip

Now, please write a detailed and coherent description of audio information in English.\\
Include: \\
- all primary sound events\\
- who and what makes the sounds\\
- where the audio happens

\medskip

Instructions:\\
- Refer to the information provided above, but note that some parts may be inaccurate and can be safely ignored.\\
- Focus primarily on meaningful audio content, while ignoring irrelevant background noise.\\
- The video caption may reference audible events; however, note that not all visible objects necessarily produce sounds.\\
- Include sounds that may originate from off-screen or partially visible sources.\\
- Maintain precise temporal alignment between visual and auditory elements, especially for distinct sound events, and narrate in chronological order.\\
- Provide a clear, detailed, and objective narration without subjective or inferential expressions. Do not use brackets, headings, or explicit labels—integrate all details smoothly into a continuous narrative.

\medskip

The following are some examples for reference: \\
- In the park, children are running and cheering, and occasionally there are dogs barking and birds singing in the distance.\\
- Deep in the forest, footsteps trampled over fallen leaves, birds chirped intermittently, and there was the occasional sound of branches breaking.

\end{casestudy}

\begin{casestudy}{Segment-level Audio-Visual Caption Generation Prompt}

I will give you some information from a video and its audio, audio is extracted from the video.

\medskip

For the previous segment:\\
The audio-visual caption is:\\
\{\textit{audio-visual caption in previous segment}\}

\medskip

For the current segment from\{\textit{timestamp[0]}\} to \{\textit{timestamp[1]}\} seconds:\\
The video caption is:\\
\{\textit{video caption in current segment}\}

\medskip

The audio caption for reference is: \\
\{\textit{audio caption in current segment}\}

\medskip

Now, please capture the contents of both vision and audio and write a detailed and coherent description for vision-audio information in English, that seamlessly integrates all visual and audio information in a balanced way.\\
Include: \\
- What is seen and what is heard\\
- Who or what is producing the sounds and appearing in the scene\\
- The environment or setting implied by both audio and visuals

\medskip

Instructions:\\
- Refer to the information provided above, but note that some parts may be inaccurate and can be safely ignored.\\
- Ensure the description contains complete and detailed visual information.\\
- Focus primarily on meaningful audio content, while ignoring irrelevant background noise.\\
- The video caption may reference audible events; however, note that not all visible objects necessarily produce sounds.\\
- Include sounds that may originate from off-screen or partially visible sources.\\
- Maintain precise temporal alignment between visual and auditory elements, and narrate in chronological order.\\
- Provide a clear, detailed, and objective narration without subjective or inferential expressions. Do not use brackets, headings, or explicit labels—integrate all details smoothly into a continuous narrative.

\medskip

The following are some examples for reference:\\
- A man reads a book in a cozy café, surrounded by the soft clatter of dishes, quiet conversation, and warm light filtering through the window.\\
- Waves crash steadily along the shore as children build sandcastles, and seagulls call overhead in the bright afternoon sun.

\end{casestudy}

\begin{casestudy}{Video-level/Global Audio-Visual Caption Generation Prompt}

I will give you some information from a video and its audio, audio is extracted from the video.

\medskip

The overall video caption for reference is:\\
\{\textit{video caption}\}

\medskip

The overall audio caption for reference is:\\
\{\textit{audio caption}\}

\medskip

The video is divided into multiple segments, and the audio-visual captions with timestamps in order are:\\
From \{\textit{timestamp[0] of segment 1}\} to \{\textit{timestamp[1]  of segment 1}\} seconds: \\
\{\textit{audio-visual caption in segment 1}\}\\
From \{\textit{timestamp[0] of segment 2}\} to \{\textit{timestamp[0] of segment 2}\} seconds: \\
\{\textit{audio-visual caption in segment 2}\}\\
…\\
From \{\textit{timestamp[0] of segment n}\} to \{\textit{timestamp[0] of segment n}\} seconds: \\
\{\textit{audio-visual caption in segment n}\}

\medskip

Now, please capture the contents of both vision and audio and write a detailed and coherent description the vision-audio information in English that seamlessly integrates all content and changes of visual and audio over time in a balanced way.\\
Include: \\
- What is seen and what is heard\\
- Who or what is producing the sounds and appearing in the scene\\
- The environment or setting implied by both audio and vision.

\medskip

Instructions:\\
- Refer to the information provided above, but note that some parts may be inaccurate and can be safely ignored.\\
- Ensure the description contains complete and detailed visual information.\\
- Focus primarily on meaningful audio content, while ignoring irrelevant background noise.\\
- The video caption may reference audible events; however, note that not all visible objects necessarily produce sounds.\\
- Include sounds that may originate from off-screen or partially visible sources.\\
- Maintain precise temporal alignment between visual and auditory elements, and narrate in chronological order.\\
- Provide a clear, detailed, and objective narration without subjective or inferential expressions. Do not use brackets, headings, or explicit labels—integrate all details smoothly into a continuous narrative.

\medskip

The following are some examples for reference:\\
- A man reads a book in a cozy café, surrounded by the soft clatter of dishes, quiet conversation, and warm light filtering through the window.\\
- Waves crash steadily along the shore as children build sandcastles, and seagulls call overhead in the bright sun.

\end{casestudy}

\begin{casestudy}{Audio-Visual Caption Verification Prompt}

Below is an existing audio-visual caption of video from \{\textit{timestamps[0]}\} to \{\textit{timestamps[1]}\} seconds to be evaluated:\\
\{\textit{audio-visual caption}\}

\medskip

Please analyze the video (with audio + visual streams) and evaluate the quality of this audio-visual caption.  \\
Identify all problems, including inaccuracies, omissions, and hallucinated content. 

\medskip

Your output should include:\\
1. **Error list**: for each error, provide the problem type, and a brief explanation.  \\
2. **Overall quality score**: a number from 1 to 5 (use “1/5, 2/5, … 5/5”) that reflects the overall accuracy, completeness and richness of details in both audio and visual aspects of the video.\\
3. **Rewritten caption**: Produce a corrected version of the caption, **only modifying** the parts that are wrong, remove meaningless background noise descriptions, and reuse all the originally correct sentences. 

\medskip

output format:
\begin{verbatim}
{
    "errors": [
        {
            "type": "[Problem Type]",
            "explanation": "[Explanation]"
        }],
    "score": [Overall quality score],
    "rewritten caption": "[Rewritten caption]"
}
\end{verbatim}

\end{casestudy}

\begin{casestudy}{QA Generation Prompt (take object-targeted question for example)}

You are an expert vision-audio question-answer dataset creator.\\
I give you the information for the video: 

\medskip

Audio-Visual caption \{\textit{timestamp}\}: \\
\{\textit{audio-visual caption}\}

\medskip

Given the above captions, your task is to generate a question-answer pair for the video:\\
Create a question that specifically targets the objects mentioned in the caption. The question should focus on identifying the objects, describing their characteristics, or understanding their roles in the video.

\medskip

The generated question is one open-ended question in English based on the given video caption. Your output should follow these rules:\\
- Encourage questions that require a descriptive or explanatory answer.
- The question should be natural and relevant and provide a clear and correct answer to the question.\\
- Ensure a question that can only be answered by combining audio and visual information from the video. \\
- The generated question-answer pairs must be grounded solely in the objective facts presented in the video's captions, without relying on any external knowledge or speculative inference.\\
- Provide the reasoning someone used to arrive at the correct answer to the question, expressed as a result of the video alone, without any captions.
\begin{adjustwidth}{1em}{0pt}
1. Question Decomposition: Break the question into smaller, manageable sub-questions without explicitly referencing or revealing elements that appear in the correct answer.\\
2. Temporal Grounding: Identify the relevant moments (timestamps or segments) in the video for each sub-question. Track how events unfold over time and highlight when key actions or changes occur.\\
3. Visual Perception: At the grounded moments, analyze visual through steps such as object detection, scene understanding, and action recognition.\\
4. Audio Perception: At the grounded moments, analyze audio through steps such as sound event detection, speech understanding, and music analysis.\\
5. Multimodal Reasoning and Answer Synthesis: Integrate visual and audio cues at the relevant timepoints to cross-validate observations, understand context, and derive the final answer clearly.\\
\end{adjustwidth}

\medskip

Output format (JSON):
\begin{verbatim}
{
    "question": "<your question>",
    "answer": "<your answer>",
    "reasoning": {
            "question decomposition": "...",
            "temporal grounding": "...",
            "visual perception": "...",
            "audio perception": "...",
            "multimodal reasoning and answer synthesis": "...",
        }
}
\end{verbatim}

\end{casestudy}

\onecolumn

\subsection{Data Visualization}
\begin{figure*}[htbp]
    \centering
    \includegraphics[width=0.96\linewidth]{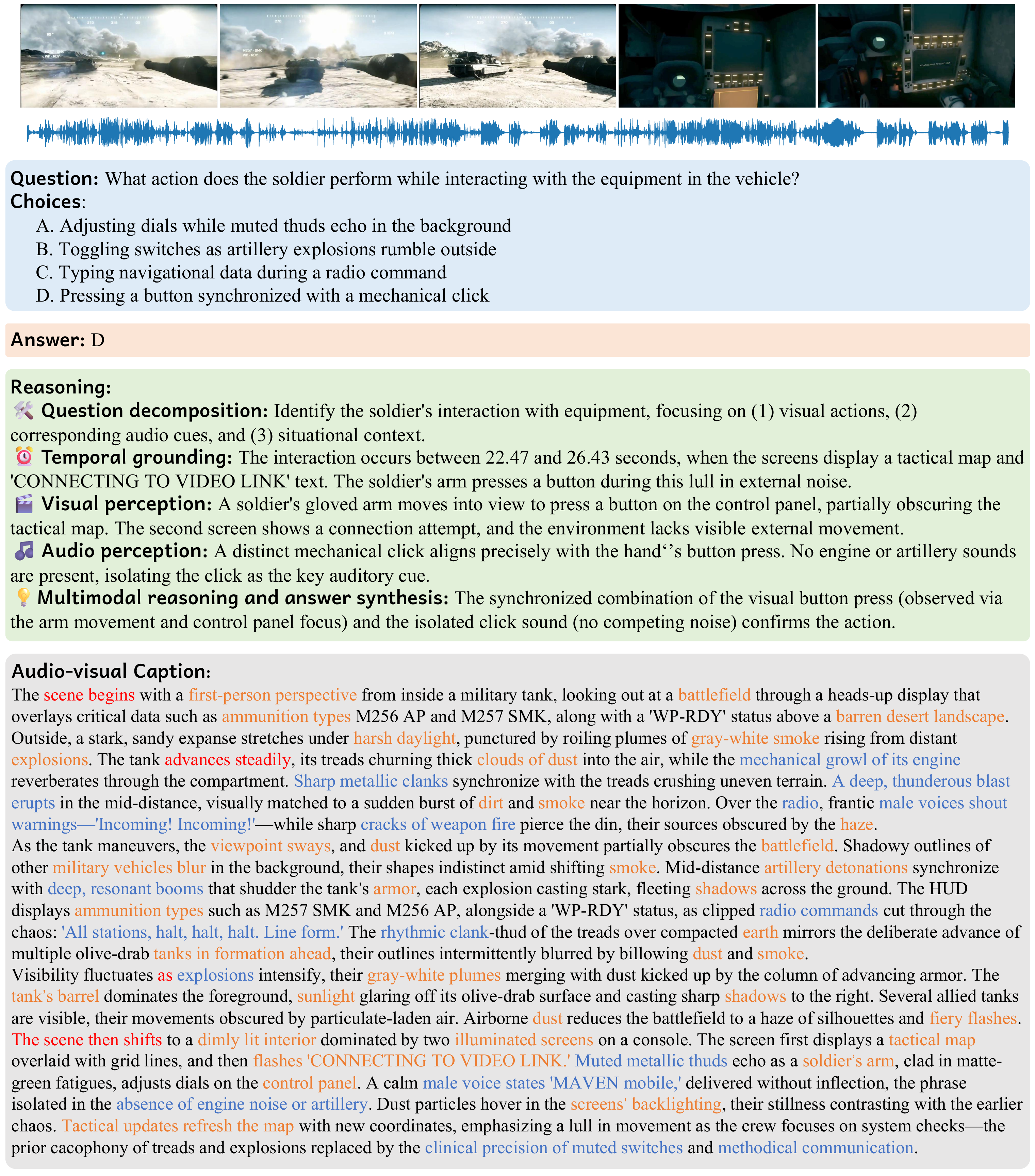}
    \caption{Example of a multiple-choice problem. Time-related content is shown in red, and visual and audio cues in orange and blue.}
    \label{fig:case1}
\end{figure*}

\begin{figure*}[htbp]
    \centering
    \includegraphics[width=0.96\linewidth]{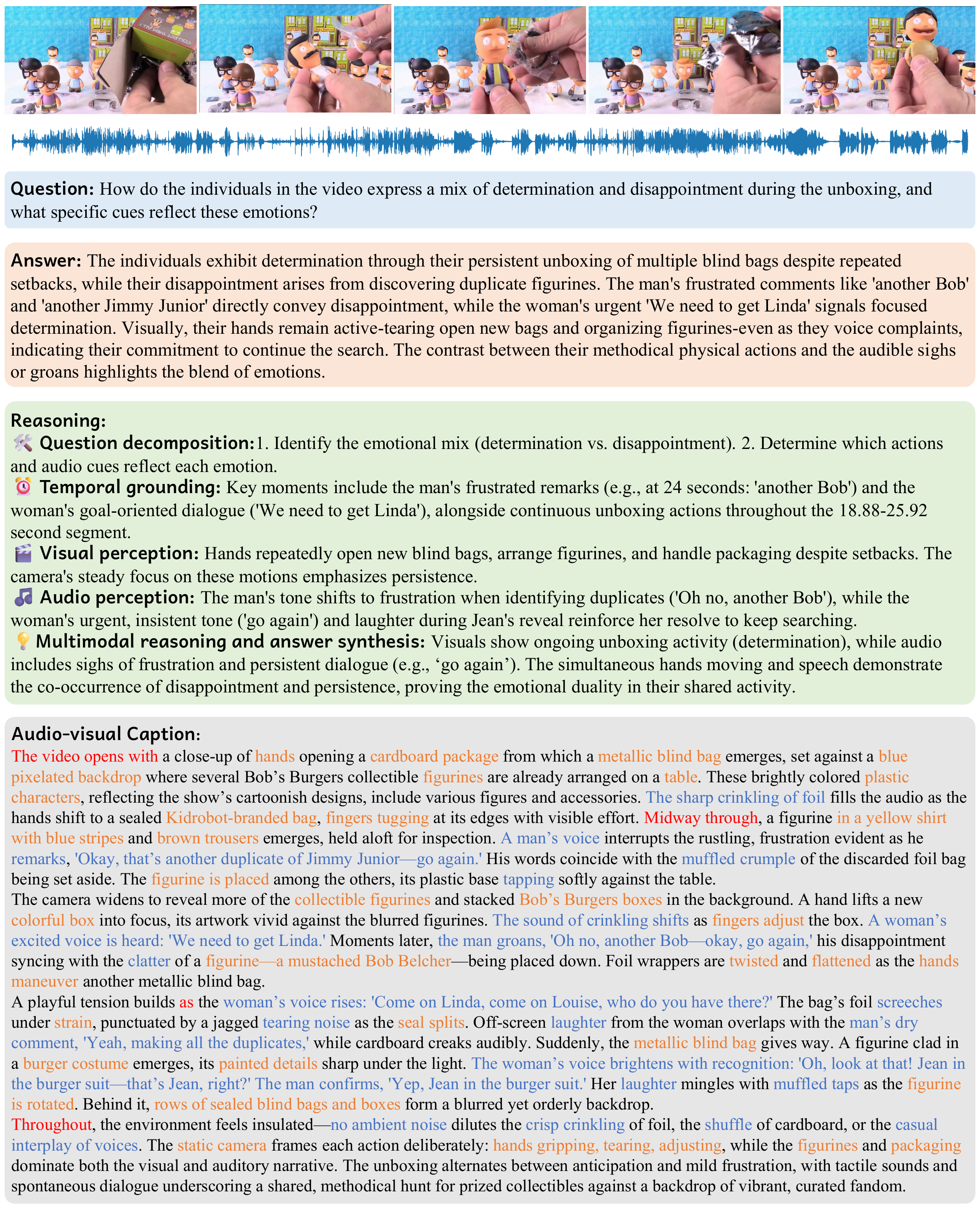}
    \caption{Example of open-ended problem. Time-related content is shown in red, and visual and audio cues in orange and blue.}
    \label{fig:case2}
\end{figure*}

In this section, we present examples in Figure~\ref{fig:case1} and Figure~\ref{fig:case2} that showcase the generated captions and QA pairs, illustrating how our pipeline produces coherent, fine-grained descriptions and reasoning annotations for each video.


\subsection{Results Visualization}
\begin{figure*}[htbp]
    \centering
    \includegraphics[width=0.96\linewidth]{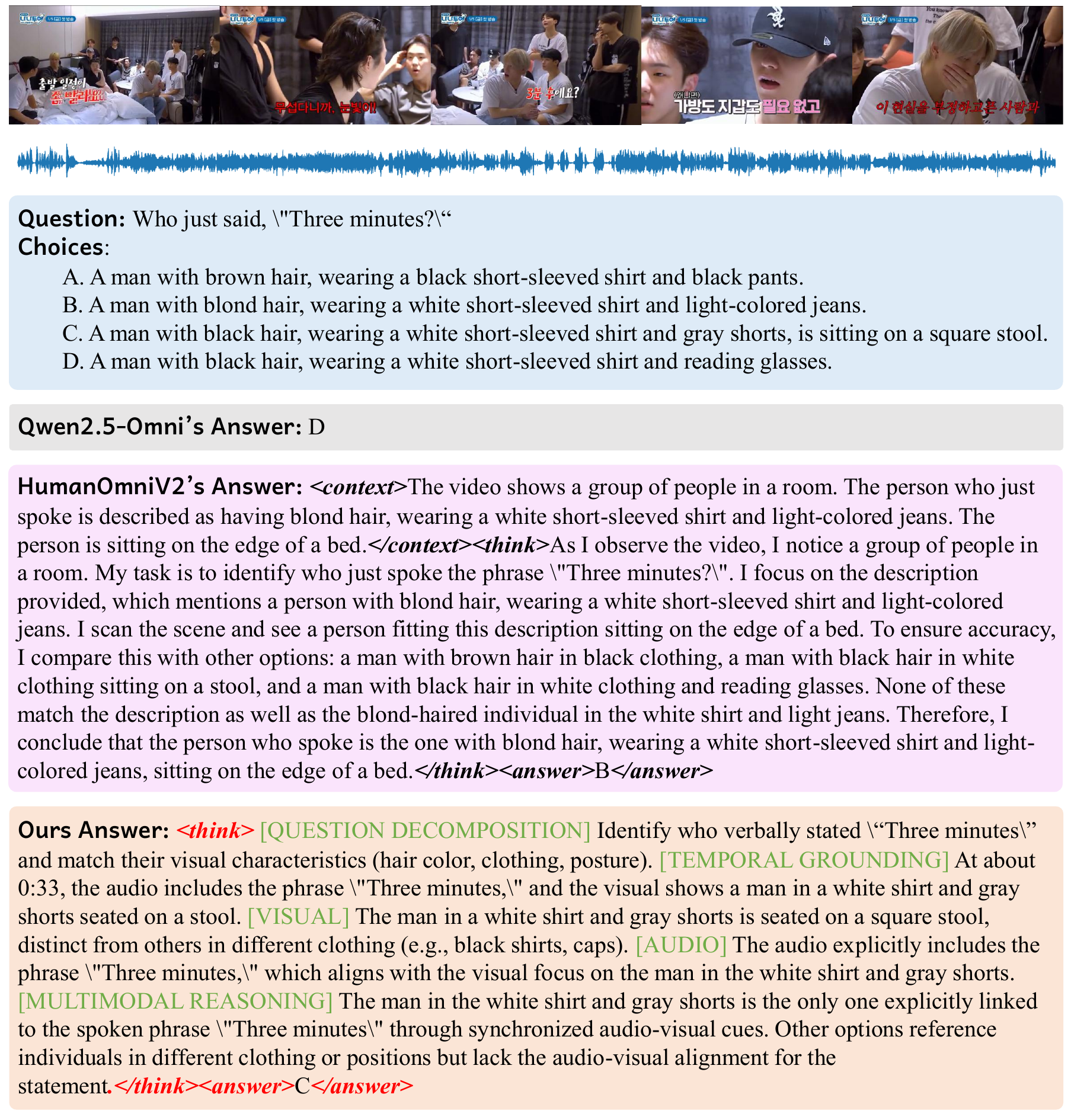}
    \caption{Example of QA results. The parts highlighted in \textcolor{red}{\textit{\textbf{{$<$red$>$}}}} and \textcolor{green1}{[green]} indicate thinking-answer process and reasoning steps.}
    \label{fig:qa_case1}
\end{figure*}

\begin{figure*}[htbp]
    \centering
    \includegraphics[width=0.96\linewidth]{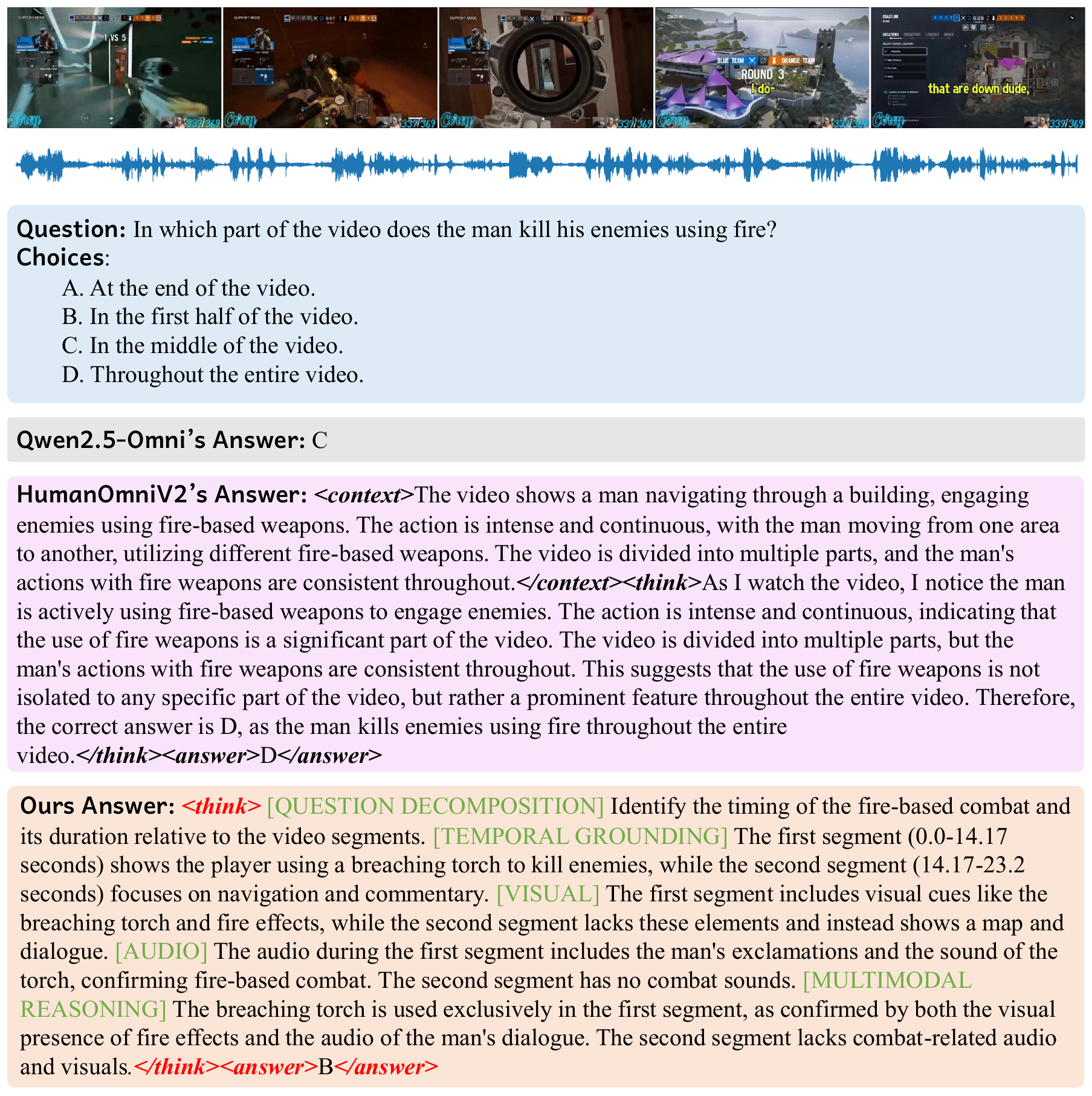}
    \caption{Example of QA results. The parts highlighted in \textcolor{red}{\textit{\textbf{{$<$red$>$}}}} and \textcolor{green1}{[green]} indicate thinking-answer process and reasoning steps.}
    \label{fig:qa_case2}
\end{figure*}

In this section, we present our model’s outputs on benchmark QA tasks in Figure~\ref{fig:qa_case1} and Figure~\ref{fig:qa_case2}, alongside results from Qwen2.5-Omni~\cite{xu2025qwen2} and HumanOmniV2~\cite{yang2025humanomniv2} for comparison.

\vspace{-5.0em}


\end{document}